\documentclass[letterpaper, 10 pt, journal, twoside]{IEEEtran}
\usepackage{amsmath,amsfonts,amssymb}
\usepackage{algorithm,algorithmic}
\usepackage{array}
\usepackage[caption=false,font=scriptsize,labelfont=sf,textfont=sf]{subfig}
\usepackage{textcomp}
\usepackage{stfloats}
\usepackage{verbatim}
\usepackage{graphicx}

\usepackage{cite}
\usepackage{gensymb}
\usepackage{booktabs}
\usepackage{float}
\usepackage{tikz}
\usepackage{mathrsfs}
\usepackage{bm}
\usepackage{makecell}
\usepackage{color,xcolor}
\usepackage{babel}
\usepackage{booktabs}
\usepackage{threeparttable}
\usepackage{url}

\def\transpose#1{{#1}^{\mathrm{T}}}

\begin{document}

\title{
Coarse-to-fine Hybrid 3D Mapping System with Co-calibrated Omnidirectional Camera and Non-repetitive LiDAR
}

\author{Ziliang Miao$^{1}$, Buwei He$^{1}$, Wenya Xie$^{1}$, Wenquan Zhao$^{1}$, Xiao Huang$^{1}$, Jian Bai$^{2}$, and Xiaoping Hong$^{1}$
\thanks{Manuscript received: Nov. 21, 2022; Revised Jan. 19, 2023; Accepted Jan. 28, 2023.}
\thanks{This paper was recommended for publication by Editor Javier Civera upon evaluation of the Associate Editor and Reviewers' comments.
This work was supported by Shenzhen Science and Technology Project (JSGG20211029095803004, JSGG20201103100401004) and SUSTech startup fund. (Ziliang Miao and Buwei He contributed equally to this work; Corresponding author: Xiaoping Hong)}
\thanks{$^{1} $These authors are with School of System Design and Intelligent Manufacturing (SDIM), Southern University of Science and Technology (SUSTech), China
        {\tt\footnotesize miaozl2019@mail.sustech.edu.cn, hebw2019@mail.sustech.edu.cn, hongxp@sustech.edu.cn}}%
\thanks{$^{2} $Jian Bai is with State Key Laboratory of Modern Optical Instrumentation, Zhejiang University, China}%
\thanks{Digital Object Identifier (DOI): see top of this page.}
}

\markboth{IEEE Robotics and Automation Letters. Preprint Version. January, 2023}{Miao \MakeLowercase{\textit{et al}}: Coarse-to-fine Hybrid 3D Mapping System with Co-calibrated Omnidirectional Camera and Non-repetitive LiDAR}

\maketitle

\begin{abstract}
This paper presents a novel 3D mapping robot with an omnidirectional field-of-view (FoV) sensor suite composed of a non-repetitive LiDAR and an omnidirectional camera. Thanks to the non-repetitive scanning nature of the LiDAR, an automatic targetless co-calibration method is proposed to simultaneously calibrate the intrinsic parameters for the omnidirectional camera and the extrinsic parameters for the camera and LiDAR, which is crucial for the required step in bringing color and texture information to the point clouds in surveying and mapping tasks. Comparisons and analyses are made to target-based intrinsic calibration and mutual information (MI)-based extrinsic calibration, respectively. With this co-calibrated sensor suite, the hybrid mapping robot integrates both the odometry-based mapping mode and stationary mapping mode. Meanwhile, we proposed a new workflow to achieve coarse-to-fine mapping, including efficient and coarse mapping in a global environment with odometry-based mapping mode; planning for viewpoints in the region-of-interest (ROI) based on the coarse map (relies on the previous work~\cite{JFR_P4S}); navigating to each viewpoint and performing finer and more precise stationary scanning and mapping of the ROI. The fine map is stitched with the global coarse map, which provides a more efficient and precise result than the conventional stationary approaches and the emerging odometry-based approaches, respectively.
\end{abstract}


\section{Introduction}
\begin{figure}[ht]
    \centering
     \subfloat[]{
        \includegraphics[width = 0.4\linewidth]{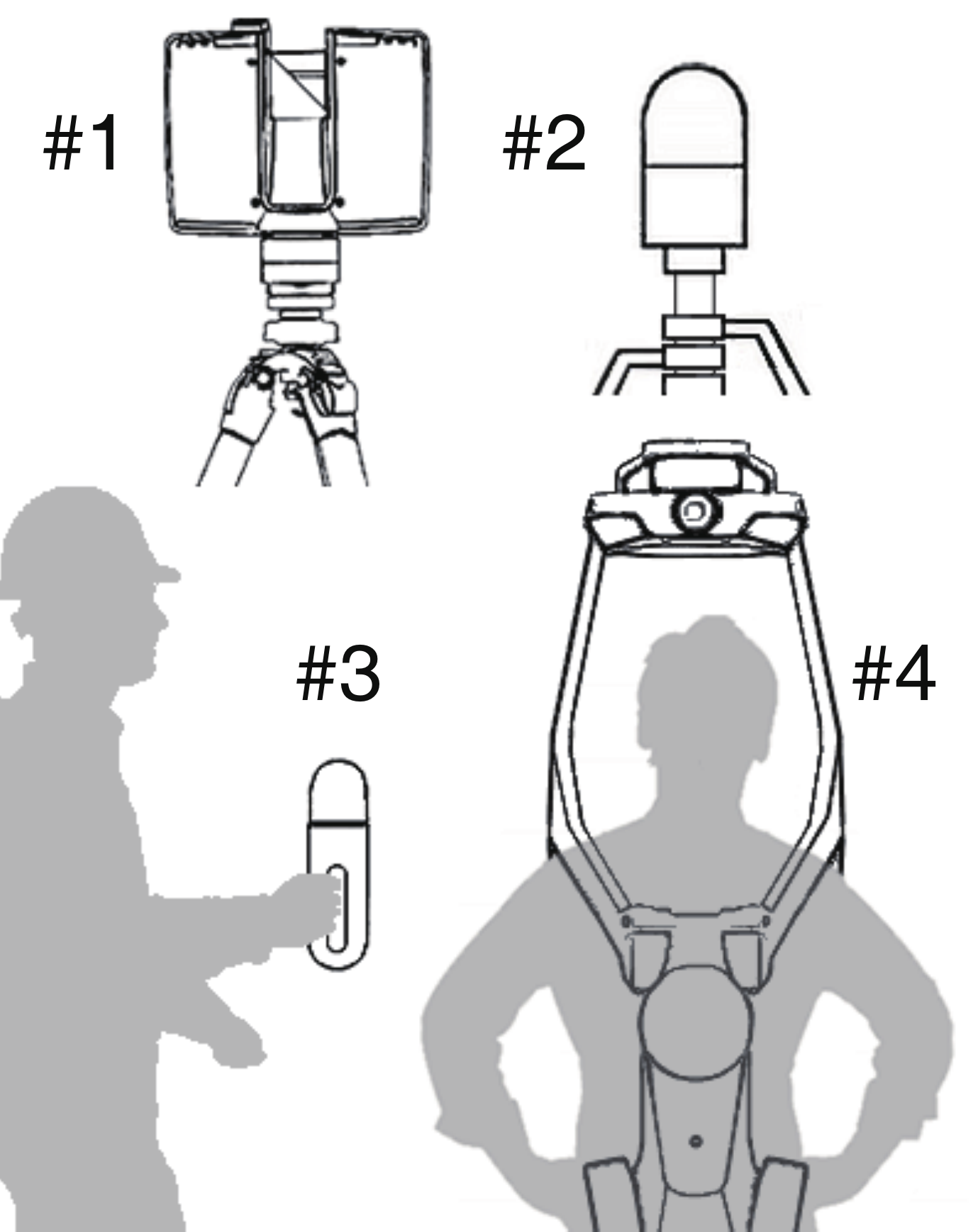} %
        \label{system_a}
    } %
    \subfloat[]{
        \includegraphics[width = 0.5\linewidth]{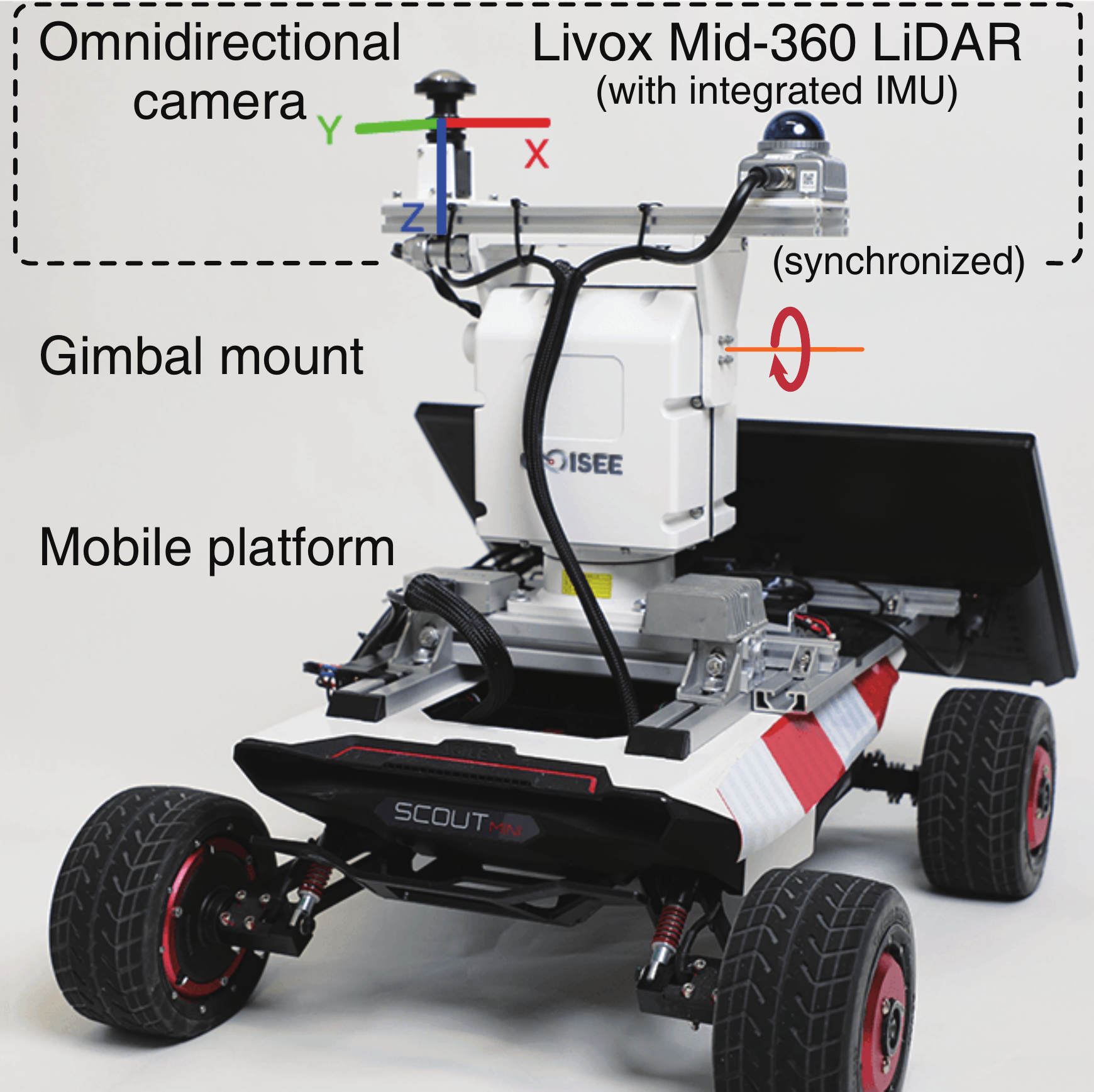} %
        \label{system_b}
    } %
    \caption{3D mapping systems: (a) the current TLS (\#1 FARO Focus Premium, \#2 LEICA BLK360) and MLS (\#3 LEICA BLK2GO, \#4 NavVis VLX) systems; (b) the proposed hybrid mapping robotic system.}
    \label{system}
\end{figure}

\IEEEPARstart{T}{hree-dimensional} scanning (obtain the raw points) and mapping (register or stitch the points into a point cloud map) are becoming increasingly important in robotics~\cite{visual-lidar-slam}, digital construction~\cite{comparison_indoor_scanning}, and virtual reality~\cite{r3live}, where digitization of the physical 3D space could provide tremendous insights in modeling, planning, management, optimization, and quality assurance. Photogrammetry has been developed to capture the 3D world. However, its application has been limited in aviation settings where accurate GPS RTK signals are required. Recently, the need for large-scale mapping of building environments has been rising, mainly due to the requirements from Building Information Modeling (BIM) systems. Thanks to the availability of emerging 3D robotic LiDAR sensors~\cite{velodyne, retina}, Mobile Laser Scanner (MLS) systems are increasingly adopted~\cite{indoor_mapping_review} (Fig.~\ref{system_a}, \#3 and \#4), where point clouds from these sensors could be registered to the global frame through sensor motion estimation (i.e., odometry) at each instance. However, due to the movement nature, such approaches largely depend on estimations of temporal characteristics such as translation and rotation, or spatial characteristics such as sensor FoV and landmark coverages. The results vary from scan to scan with no guarantee of precision. Hence, a more robust and precise method is desired.

On the other hand, the traditional Terrestrial Laser Scanner (TLS) has been employed in many precision-stringent applications (Fig.~\ref{system_a}, \#1 and \#2). The TLS-based stationary mapping is usually inefficient (due to the accurate but slow laser rotation) but could provide precise results. Viewpoints (also known as stationary scanning locations) need to be carefully planned to ensure the spatial coverage and enough overlapping regions of adjacent viewpoints to make accurate point cloud stitching~\cite{combined_SLAM_TLS}, but on the other hand, as fewer as possible to reduce scanning time and cost. The planning for viewpoints largely relies on the overall layout of the scene, which has been done by human experience so far~\cite{P4S_review}.

Combining the strength from both worlds would be ideal in large-scale 3D mapping applications. As shown in Fig.~\ref{system_b}, the proposed hybrid mapping robot is developed carrying a gimbal mount and a novel sensor suite consisting of an omnidirectional non-repetitive Livox Mid-360 LiDAR\footnote{The authors gratefully acknowledge Livox Technology for the equipment support.} and an omnidirectional camera. The sensors' FoV and the non-repetitive scanning nature are shown in Fig.~\ref{sensor_a}. In the odometry-based mapping mode, the sensor suite is kept horizontal by fixing the gimbal mount to coarsely and efficiently map the entire space with the mobile platform. Based on the coarse map, a few viewpoints are planned for the stationary mapping of targeted ROIs. In the stationary mapping mode, the robot will navigate and stay still at each viewpoint, performing $360\degree \times 300\degree$ scanning by traversing the vertical FoV through the gimbal mount. These precise scans are registered with each other and then stitched with the pre-generated coarse map forming a global map with fine ROIs.

The main contributions of this work are as follows:
\begin{itemize}
  \item [1)]
  The first hybrid 3D mapping robot system that integrates odometry-based and stationary mapping modes is proposed. The consistency of point clouds in two modes can be guaranteed with the single omnidirectional non-repetitive Livox Mid-360 LiDAR.
  \item [2)]
  An omnidirectional camera is introduced in the proposed system to complement the omnidirectional LiDAR. A novel automatic targetless co-calibration method is proposed to simultaneously calibrate the intrinsic parameters and the extrinsic parameters.
  \item [3)]
  An automated coarse-to-fine hybrid mapping workflow is demonstrated, including odometry-based coarse mapping in the global environment, planning for the viewpoints in the ROIs, and finer stationary mapping at viewpoints. The entire project is open-sourced on GitHub\footnote{https://github.com/ZiliangMiao/Hybrid\_Mapping\_Cocalibration.git} to aid the development of this emerging field.
\end{itemize}

\section{Related Works}
\subsection{Mapping Solutions}
3D mapping solutions are of great interest in many emerging fields~\cite{comparison_indoor_scanning}. TLS-based and MLS-based approaches are commonly adopted. 

The traditional TLS-based approach uses a heavy-duty single-laser scanner and traverses the entire FoV through step-wise rotations about the horizontal and vertical axes. It provides sufficiently dense points with good precision. However, this method is slow and laborious. It has to be repeated on many viewpoints, which need to be chosen wisely because a lack of viewpoints will cause missing information in the desired ROI, while the excess of viewpoints will lead to longer scanning hours and poorer efficiency. Currently, viewpoints planning relies on human intuition or experiences, making it challenging to plan effectively in large and complex working environments like the construction scenes~\cite{P4S_review}.

On the contrary, the MLS-based approach provides real-time scanning and mapping results as the LiDAR moves. The current MLS devices are classified by their usage configurations, such as handheld (Fig.~\ref{system_a}, \#3), backpack (Fig.~\ref{system_a}, \#4), and trolley. Most of these mobile systems rely on conventional LiDARs (16, 32, or 64 lines) and construct the 3D map by registering the point cloud with LiDAR odometry or LiDAR-IMU odometry. Such mobile systems greatly speed up the mapping process without planning for viewpoints. However, it cannot replace the TLS-based approaches due to insufficient mapping precision and sparse point clouds~\cite{comparison_indoor_scanning}. The repetitive scanning nature of mechanical LiDAR is unsuitable for stationary scanning due to limited FoV coverage (20\% coverage for 32-line LiDAR). Therefore, the indispensable motion for more coverage will cause errors in pose estimation, which are accumulated throughout the process, limiting the usage in high-precision applications.

Both TLS-based and MLS-based approaches have their unique advantages and drawbacks. It is desired to devise a mechanism to combine both modes. For example, a combination of TLS and MLS is used to solve the registration problem between non-overlapping spaces~\cite{combined_SLAM_TLS} or use TLS scans as references to MLS mapping registration to achieve low mapping errors~\cite{slam-aided}. Moreover, MLS is also used to provide a 3D map to solve the viewpoints planning problem of TLS~\cite{P4S_review}. However, all these methods are based on heterogeneous sensors for different modes, with different synchronization, data structure, and protocols, which are difficult to construct a one-stop mapping robot with a streamlined and automated workflow.

The unique non-repetitive scanning nature of the Livox LiDAR provides a combination of an instantaneous high density at a short time interval for odometry (with effective point density as 32-line LiDAR within 0.1 seconds) and an image-level resolution at relatively long time intervals for scanning (within 3 seconds, as shown in Fig.~\ref{sensor_b}), which makes it surprisingly suitable for such hybrid working mechanism. The feature provides sufficiently good performance in odometry scenarios~\cite{fastlio2} and a dense FoV coverage for image-like feature processing~\cite{retina, camvox, voxel}. In this paper, the two working modes are integrated into the same robot, ensuring overall mapping efficiency and precision with an automated coarse-to-fine hybrid mapping workflow.
\subsection{Calibration Methods}
In addition to LiDAR, cameras are usually required in 3D mapping systems to give an overview of the mapped environment~\cite{review_of_mobile_mapping}. Cameras could provide high-quality geometric, color, and texture information~\cite{camera-LiDAR_integration}, which enables further modeling and rendering~\cite{point_cloud_rendering} of the point clouds and permits tasks in object detection, segmentation, and classification~\cite{multi-sensoral_mobile_mapping}. Meanwhile, for autonomous navigation, the camera is also vital to visual-LiDAR odometry through sensor fusion~\cite{r3live}. All these functions would rely on the accurate calibration of the intrinsic parameters of the camera and extrinsic parameters between the cameras and LiDAR~\cite{camera-LiDAR_integration}. 

Traditionally, multiple cameras are usually required to be complementary to the omnidirectional FoV of LiDAR. This work employs an omnidirectional camera over the traditional multi-camera vision to avoid bulky construction, high cost, shutter synchronization, and cascaded extrinsic calibrations. The intrinsic and extrinsic parameters of this novel omnidirectional sensor suite are essentially needed.

The intrinsic parameters of the omnidirectional camera must be well calibrated since these types usually possess much larger and more complex distortions than pin-hole cameras~\cite{generic}. In~\cite{ocamcalib,japan,generic}, higher-order polynomial-based intrinsic models are introduced with many degrees of freedom to obtain satisfactory results. A popular OcamCalib toolbox based on the checkerboard is provided~\cite{ocamcalib}. These methods could be susceptible to over-fitting with high-order polynomials and often require evenly distributed artificial targets and dense features across the entire space. Typically, these calibration processes are manual and could lead to tedious procedures with a large margin of error. Additionally, the omnidirectional camera in our work is constructed with a refractive-reflective geometry to capture a ring-like FoV beyond $180\degree$. This construction makes intrinsic calibration even more difficult. An accurate, automatic, and targetless calibration method is desired.

The extrinsic calibration method between the omnidirectional camera and LiDAR has only been explored in~\cite{fisheye_extrinsic} using edge correspondence to match point clouds and images. The bearing angle images highlight the edge features, which are manually positioned. Targetless extrinsic calibration methods for monocular cameras and LiDAR have been developed recently. With the non-repetitive LiDARs, CamVox~\cite{camvox} could project the image-like LiDAR point clouds onto the camera image plane and extract edge pixels using the grayscale images based on reflectivity and depth. The method proposed in~\cite{voxel} uses voxels to extract the edge points in 3D space and classifies the edges based on depth continuity. Both methods work well with conventional pin-hole cameras and need to be extended toward the omnidirectional cameras with significantly larger distortions. An additional targetless extrinsic calibration method employing mutual information (MI) is also developed~\cite{mutual}, which maximizes the intensity correlations of LiDAR and camera. However, the misrepresented information caused by lighting conditions, surface reflection properties, and spectral reflectance disagreement could result in worse calibration than the edge-based methods.

In the proposed targetless co-calibration method, the high-resolution dense point cloud of the non-repetitive scanning LiDAR gives abundant and ground-truth-level features, which eliminates the artificial targets and manual involvement and reduces the error caused by insufficient coverage and sparse features of the targets. With the co-calibration method, the intrinsic and extrinsic parameters are obtained simultaneously and can be re-calibrated fast and reliably in work scenes.

\section{Proposed System}
\subsection{Co-calibrated Omnidirectional Sensor Suite}
\begin{figure}[ht]
    \centering
    \subfloat[]{
        \includegraphics[width = 0.75\linewidth]{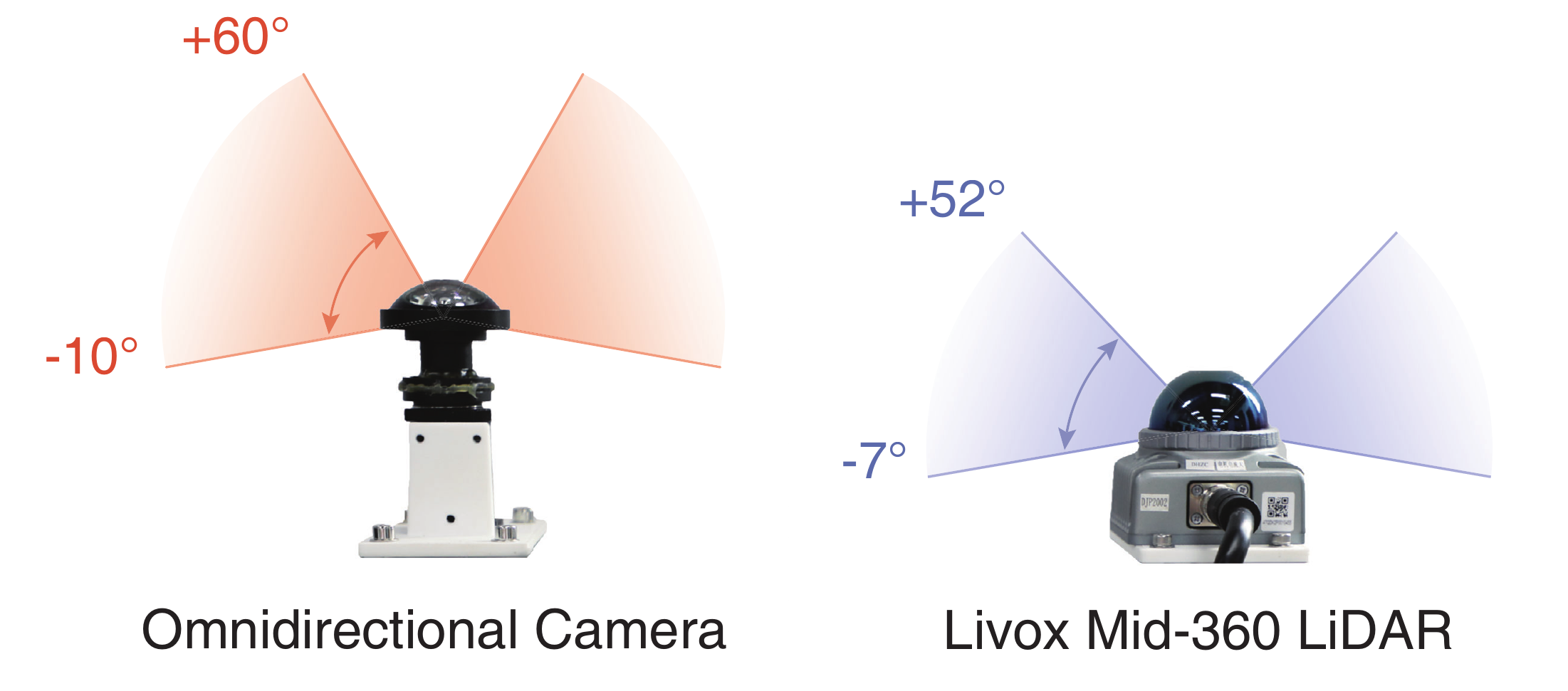}
        \label{sensor_a}
    } %
    \vfill
    \subfloat[]{
        \includegraphics[width = 0.75\linewidth]{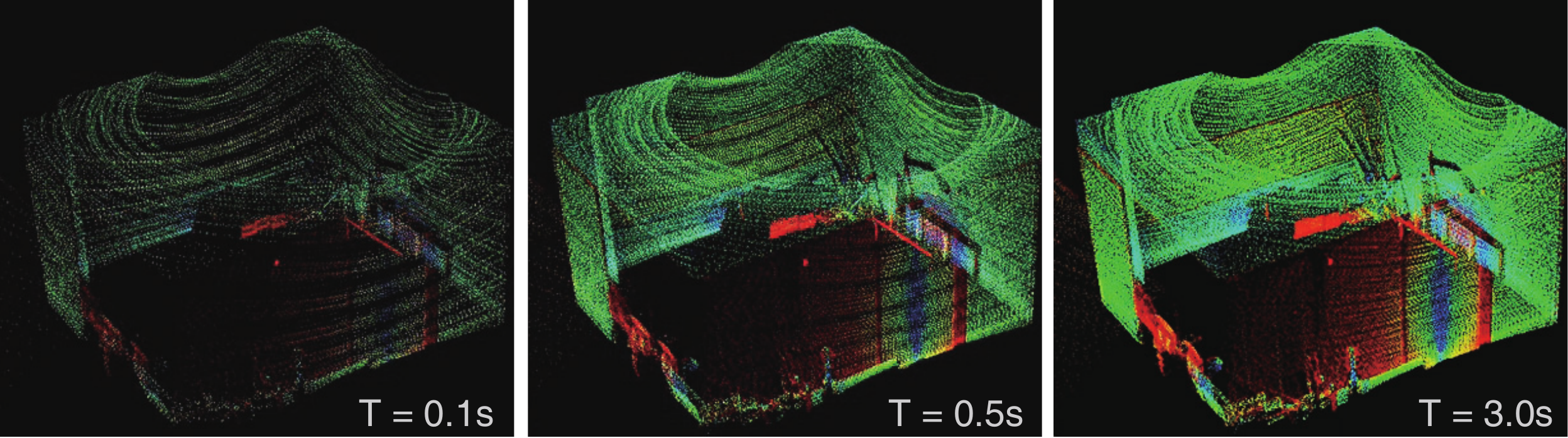}
        \label{sensor_b}
    } %
    \caption{Configuration of the sensors: (a) omnidirectional camera and Livox Mid-360 LiDAR, both on the gimbal mount; (b) point cloud accumulation over time due to the non-repetitive scanning nature of the Livox LiDAR.}
    \label{sensor}
\end{figure}

The Livox Mid-360 LiDAR has a $360\degree \times 55\degree$ FoV and features a non-repetitive scanning pattern, with increasingly denser points over time (the coverage of FoV approaches 100\%), as shown in Fig.~\ref{sensor_b}. The unique feature specifically benefits both odometry-based and stationary mapping modes. The omnidirectional camera provides color information of the surroundings and has a corresponding $360\degree \times 70\degree$ FoV (Fig.~\ref{sensor_a}). Both sensors are synchronized and are mounted on a two-axis gimbal (Fig.~\ref{system_b}) to extend the scanning FoV to $360\degree \times 300\degree$.

\vspace{-12pt}
\begin{figure}[htp]
    \centering
    \includegraphics[width = 0.95\linewidth]{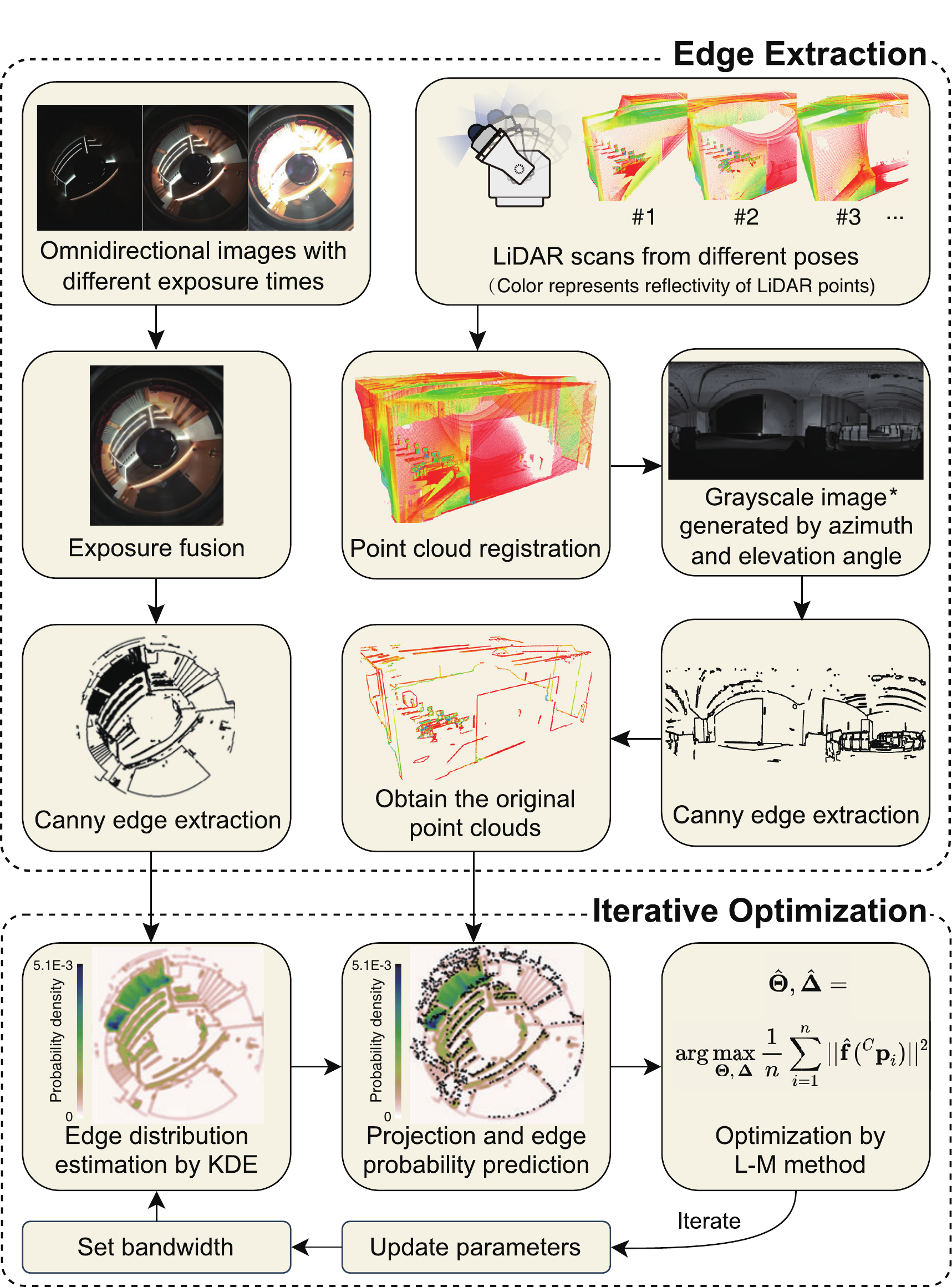}
   \caption{Proposed co-calibration process. *~The grayscale value indicates the average reflectivity of the projected LiDAR points within a pixel.}
    \label{cocalibration_process}
\end{figure}

The co-calibration simultaneously obtains the intrinsic (camera) and extrinsic (camera-LiDAR) parameters, defined respectively as $\mathbf{\Theta}\triangleq\transpose{[u_0, v_0, c, d, e, a_0, \dots, a_n]}$ and $\mathbf{\Delta}\triangleq\transpose{[\alpha, \beta, \gamma, t_x, t_y, t_z]}$, which will be introduced later. With the unique benefit of the non-repetitive scanning LiDAR, an extremely dense point cloud is always available, which provides a 3D ground truth of the environment. This high-resolution point cloud could be projected onto the 2D image plane with pixel values from LiDAR reflectivity, from which clear edge features could be extracted. To align the edges from LiDAR and the camera, the co-calibration iteratively maximizes the correspondence of projected LiDAR edge points with the omnidirectional camera edge pixels. Kernel Density Estimation (KDE) is employed to estimate the camera edge distribution with different distribution smoothness (by varying bandwidth coefficient) to obtain global optimum. The entire process of co-calibration can be divided into the following two steps (Fig.~\ref{cocalibration_process}):

\subsubsection{Edge Extraction}
Edge extractions are performed for both camera and LiDAR. For the camera, exposure fusion~\cite{exposure_fusion} is adopted to enhance the dynamic range of images to capture more details for low and high-brightness objects. Canny edge extraction~\cite{canny} is performed on the enhanced image, with edge points $\mathbb{Q} = [\bm{q}_1, \bm{q}_2 \ldots, \bm{q}_n]$. For LiDAR, since the FoV is smaller, point clouds scanned from different pitch angles are stitched together. The stitching is performed by the generalized iterative closest point (GICP) algorithm~\cite{gicp} with the initial transformation given by the state of the gimbal. The stitched point cloud with reflectivity is then projected to an image plane with the azimuthal angle and elevation angle as the coordinates, generating a grayscale image by taking the average reflectivity of the projected LiDAR points within each pixel. The Canny edge extraction is performed on this grayscale image. Uniform sampling is performed in each stage to remove the non-uniform point distribution. The edge pixels are then identified in the original 3D point cloud $\mathbb{P} = [{^L}\mathbf{P}_1, {^L}\mathbf{P}_2 \ldots, {^L}\mathbf{P}_m]$.

\subsubsection{Iterative Optimization}
The iterative optimization is performed in the omnidirectional image space. The LiDAR edge points are projected to the image coordinates through the following equations:

\begin{align}
    \begin{split}\label{projection1}
        {^C}\mathbf{P} &= {^C_L}\mathbf{T}({^L}\mathbf{P};\mathbf{\Delta}) = {^C_L}\mathbf{R}\cdot{^L}\mathbf{P} + {^C_L}\mathbf{t}, {^L}\mathbf{P} \in \mathbb{P},
    \end{split}\\
        \begin{split}\label{projection2}
        \bm{p} &= \mathbf{\Pi}({^C}\mathbf{P};\mathbf{\Theta}) = \begin{bmatrix}
                c & d \\
                e & 1
        \end{bmatrix}
        \begin{bmatrix}
            r\cos{\phi} - u_0 \\
            r\sin{\phi} - v_0
        \end{bmatrix},
    \end{split}\\
    r &= \mathbf{F}(\theta; a_0, \dots, a_n) = a_0 + a_1 \theta^1 + \ldots + a_n \theta^n, \\
    \theta &= \arccos(\frac{z}{\sqrt{x^2+y^2+z^2}}), \\
    \phi &= \arccos(\frac{x}{\sqrt{x^2+y^2}}),
\end{align}
where ${^C}\mathbf{P}$ and ${^L}\mathbf{P}$ denote the 3D point coordinates in camera and LiDAR coordinate systems, respectively, and they are related through the extrinsic transformation ${^C_L}\mathbf{T}({^L}\mathbf{P};\mathbf{\Delta})$, i.e., rotation ${^C_L}\mathbf{R}$ and translation ${^C_L}\mathbf{t}$ with the extrinsic parameters $\mathbf{\Delta}$. The symbol $\bm{p}$ denotes the location of the point in the camera image space, and $\mathbf{\Pi}({^C}\mathbf{P};\mathbf{\Theta})$ expresses the intrinsic transformation from ${^C}\mathbf{P}=\transpose{[x, y, z]}$ (3D point) to $\bm{p}$ (2D point), with the distortion correction matrix $\begin{bmatrix}
        c & d \\
        e & 1
    \end{bmatrix}$. The pixel radius $r$ from the image center $\transpose{[u_0, v_0]}$ is transformed from the elevation angles $\theta$ by a polynomial function $\mathbf{F}(\theta; a_0, \dots, a_n)$ in the camera model; $\theta$ and $\phi$ are the elevation and azimuth angle of ${^C}\mathbf{P}$ (Note the omnidirectional camera features a ring image).

To facilitate the alignment between the camera edges and the LiDAR edges, the camera edge distribution with nonparametric probability density function is constructed with the Gaussian Kernel by Kernel Density Estimation (KDE)~\cite{kde}. The optimization is based on maximizing the probabilities of the projected LiDAR edge points onto the camera edge distribution:
\begin{align}
    \hat{\mathbf{\Theta}}, \hat{\mathbf{\Delta}} &= \arg \max_{\mathbf{\Theta}, ~\mathbf{\Delta}} \frac{1}{m}\sum_{i=1}^m || \hat{\bm{f}}(\bm{p}_i; h, \mathbb{Q}) ||^2, \\
    \hat{\bm{f}}(\bm{p}_i;h,\mathbb{Q}) &= \frac{1}{n h^2} \sum_{j=1}^{n} \mathbf{K}\left(\frac{\bm{p}_i-\bm{q}_{j}}{h}\right), \\
    \mathbf{K}(\bm{x}) &= \frac{1}{\sqrt{2 \pi} \det(\mathbf{\Sigma})} e^{-\frac{1}{2} \transpose{(\bm{x} - \bm{\mu})} \mathbf{\Sigma}^{-1} (\bm{x} - \bm{\mu})}, \\
    \bm{\mu} &= \transpose{[0, 0]}, ~\mathbf{\Sigma} = \mathbf{I}_{2\times2},
\end{align}
where $h$ denotes the bandwidth of the KDE.

\begin{figure}[hp]
    \centering
    \subfloat[]{
        \includegraphics[width=0.45\linewidth]{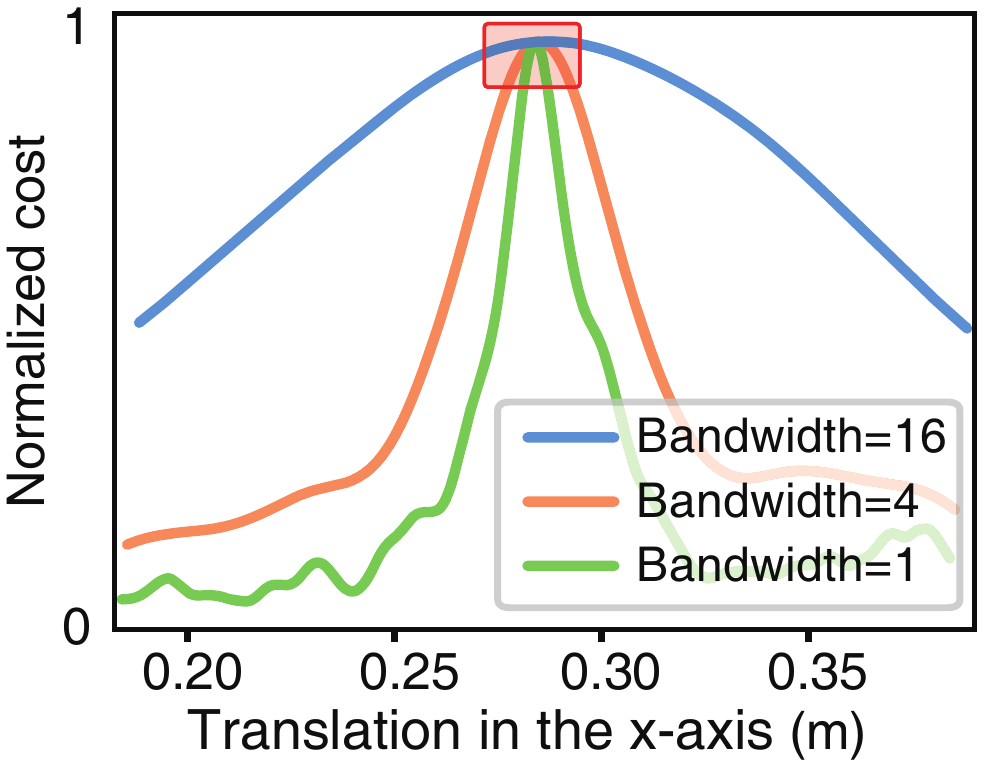}
    } %
    % \hfill
    \subfloat[]{
        \includegraphics[width=0.425\linewidth]{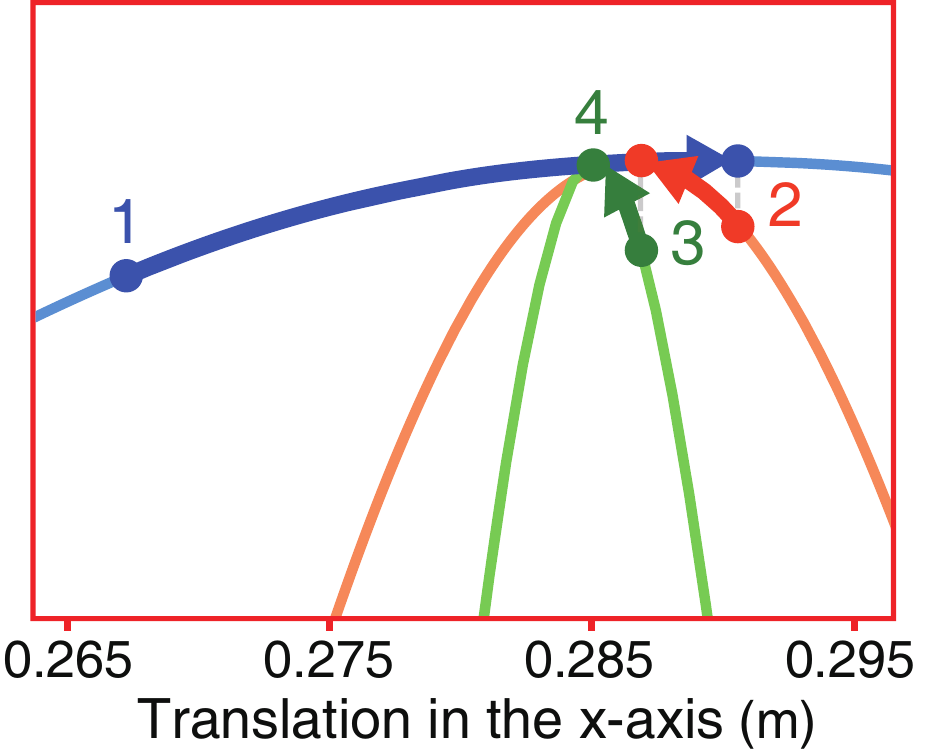}
    } %
    \caption{Iterative optimization with the reducing KDE bandwidth: (a) the normalized cost w.r.t. the translation in the x-axis under the different values of bandwidth; (b) zoom in to a sub-region of (a) to demonstrate the iterative process.}
    \label{bandwidth}
\end{figure}

Several rounds of iterative optimization with reducing bandwidth are carried out to approach the correct calibration values smoothly. At the start of the process, the bandwidth is set at a large number to get a continuous and smooth cost function, which allows the optimization to approach the optimal region quickly without many local optima. Then the bandwidth is reduced gradually to increase the gradient, ensuring a sensitive optimization around the optimum (optimization of the x-axis translation is shown in Fig.~\ref{bandwidth}).
\vspace{-12pt}

The optimization uses the Levenberg-Marquardt method implemented in Ceres-solver~\cite{ceres}. For computational efficiency, the parabolic Epanechnikov kernel $\mathbf{K}(\bm{x}) = \frac{3}{4}(1 - \transpose{\bm{x}}\bm{x})$ can be substituted for the Gaussian kernel.
\subsection{Coarse-to-fine Hybrid Mapping}
\begin{figure}[htp]
    \centering
    \includegraphics[width = 0.925\linewidth]{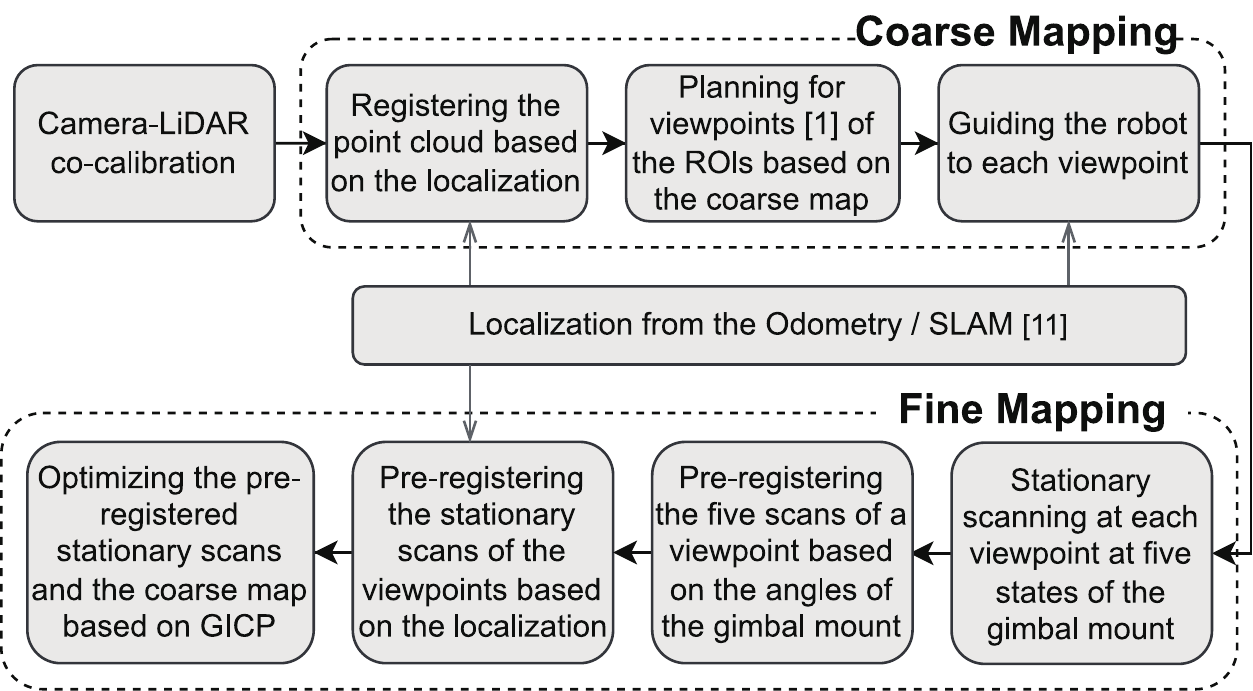}
    \caption{Proposed coarse-to-fine hybrid mapping workflow. The odometry/SLAM serves as a backbone to provide localization results.}
    \label{hybrid_workflow}
\end{figure}

 The coarse-to-fine hybrid mapping workflow is outlined in Fig.~\ref{hybrid_workflow}. With the co-calibration and synchronization, all the obtained LiDAR points are represented in both coordinates and color. Odometry/SLAM methods are used as a backbone to provide localization in both coarse and fine mapping. We used FAST-LIO (LiDAR-Inertial odometry~\cite{fastlio2}) in our current system, but the choice is not limited; other odometry/SLAM methods could be utilized as well. At the coarse mapping stage, the robot obtains the localization and motion results from the odometry, from which the scanned points are converted and registered to the global map. Based on the coarse map, a few viewpoints for stationary mapping are planned for the targeted ROIs, which is well developed in previous work by considering the constraints such as range, grazing angle, FoV, and overlap~\cite{JFR_P4S}. The robot then navigates to the generated viewpoints one-by-one through the backbone odometry/SLAM and performs the fine mapping, respectively. At each viewpoint, stationary scans are performed at several gimbal states, with overlapping FoV regions between the adjacent two states, and cover a large overall FoV ($360\degree\times300\degree$). These point clouds will be pre-registered based on the gimbal angles (as initial angles) at each viewpoint. The scans from all the viewpoints are then combined with the global coarse map based on robot localization (again provided by the LiDAR-Inertial odometry) as the initial state for optimization. Finally, the GICP~\cite{gicp} algorithm is used to optimize all the localization results and gimbal states and refine all stationary scans and the coarse map to form the fine map. Notably, we could choose either odometry or SLAM methods in the localization backbone. Although SLAM has more loop-closure functions than odometry, the final GICP optimization is accurate enough to yield a much better localization result.

\section{Experiments and Results}
\label{experiments}
\subsection{Co-calibration Results}
\label{4.0}
\begin{figure}[h]
    \centering{
    	\subfloat[]{
    	    \includegraphics[width = 0.45\linewidth]{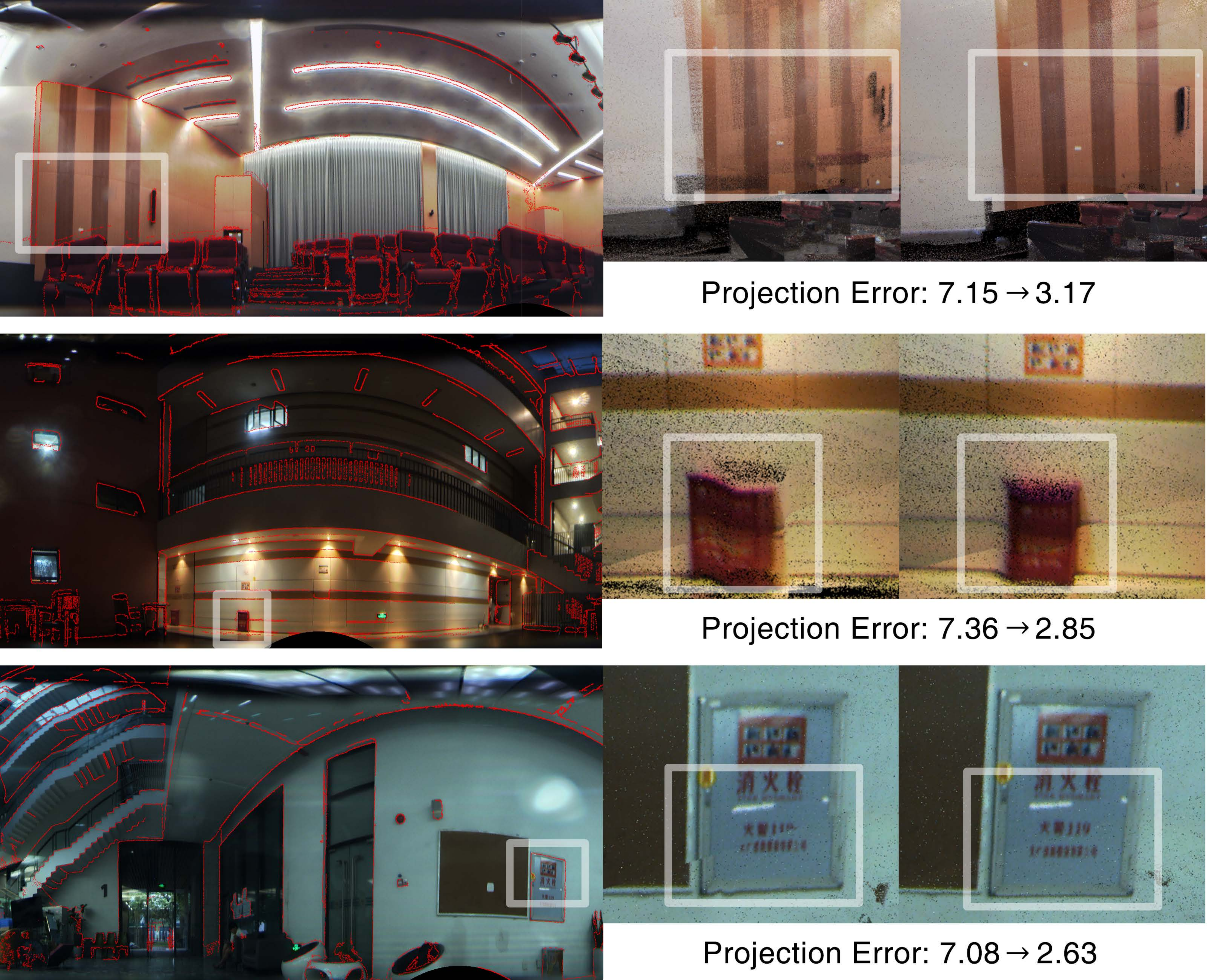}
                \label{calibration_a}
    	} %
    	\subfloat[]{
    	    \includegraphics[width = 0.45\linewidth]{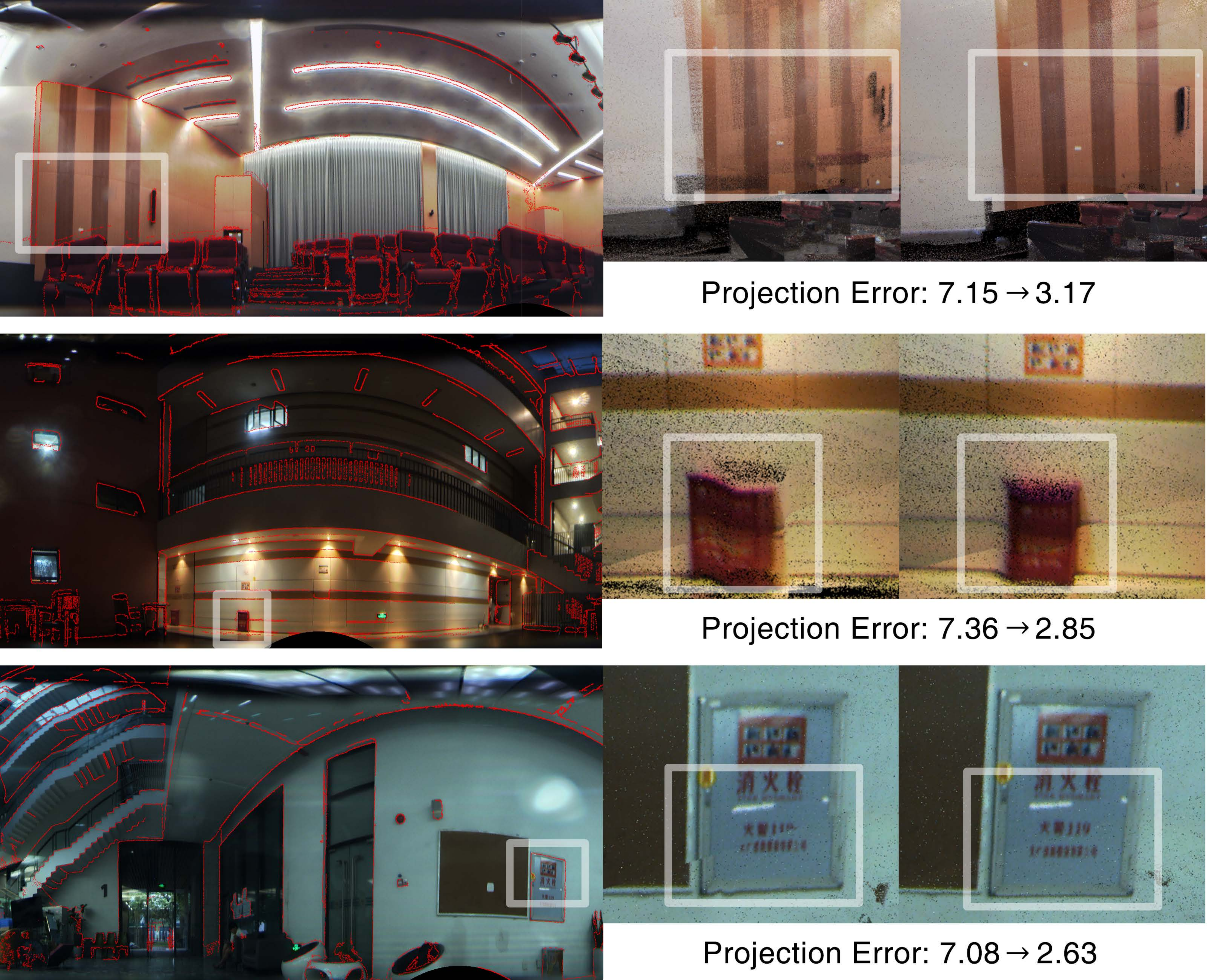}
                \label{calibration_b}
    	} %
	} %
     \caption{Co-calibration results in three scenes: (a) aligned LiDAR edge points (red) on camera images; (b) comparison of colorized point clouds before and after co-calibration with the average projection errors in pixels.}
    \label{calibration}
\end{figure}

The effectiveness of the proposed co-calibration method is demonstrated in three natural scenes, as shown in Fig.~\ref{calibration}. The projection error (in pixels) is defined as:
\begin{equation}
    e = \frac{1}{n} \sum_{i=1}^n \bm{d}(\bm{p}_i; \mathbb{Q}),
\end{equation}
where $\bm{d}$ is to calculate the distance from the LiDAR projected point $p_i$ to the nearest point in target set $\mathbb{Q}$. Note that the largest 10\% of the distances are considered outliers with no correspondences and are eliminated. Overall, the co-calibration works well in all scenes with projection errors on the order of 3 pixels or less. The colorized point clouds after co-calibration also show much better consistency, as seen in Fig.~\ref{calibration_b}.

We further compare our co-calibration results with the classical target-based intrinsic calibration~\cite{ocamcalib, improved_ocamcalib}, and the state-of-the-art MI-based extrinsic calibration~\cite{mutual}, respectively, as shown below.

\subsubsection{Analysis of the Intrinsic Results}
\begin{figure}[ht]
    \centering{
    	\subfloat[]{
    	    \includegraphics[width = 0.575\linewidth]{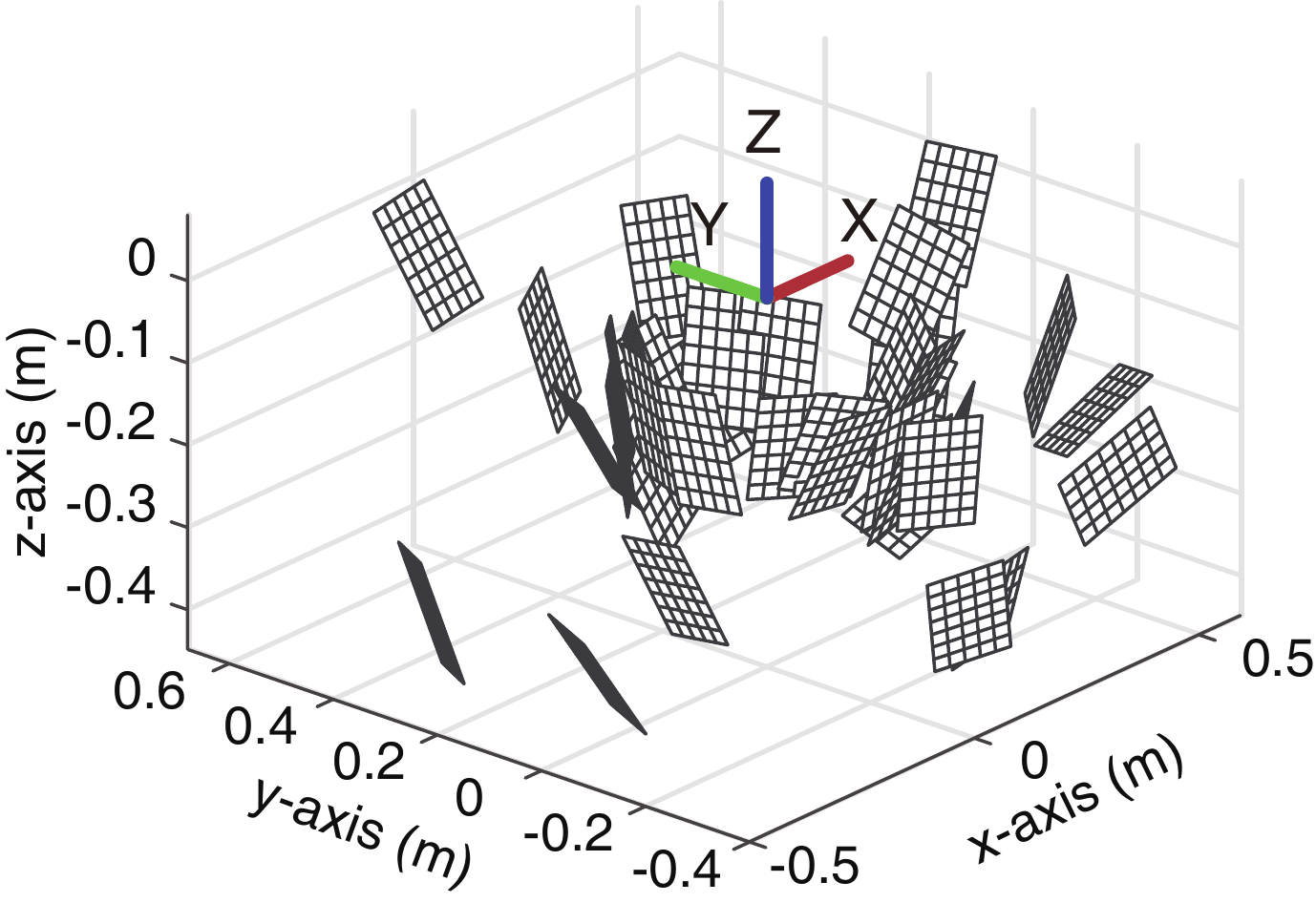}
                \label{checkerboard_a}
    	} %
    	\subfloat[]{
    	    \includegraphics[width = 0.345\linewidth]{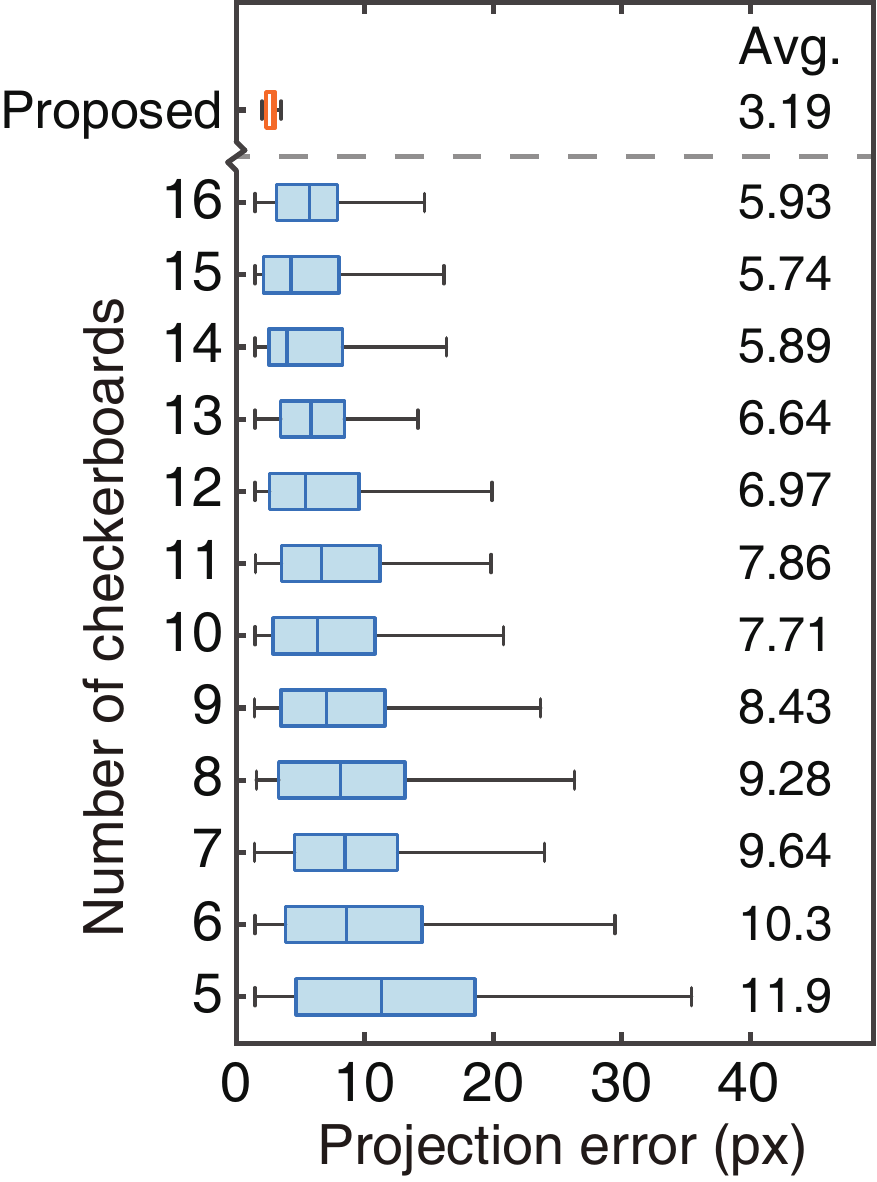}
                \label{checkerboard_b}
    	} %
	} %
    \caption{Comparison with the target-based intrinsic calibration: (a) the poses of the thirty checkerboards; (b) boxplots of projection errors of target-based calibration (blue) and the proposed co-calibration (orange).}
    \label{checkerboard}
\end{figure}

As a comparison, the target-based intrinsic calibration for omnidirectional cameras is performed~\cite{ocamcalib}. Thirty checkerboards are manually selected as a reference set (Fig.~\ref{checkerboard_a}). As the number and position of the targets affect the calibration profoundly, we evaluate the calibration result as a function of the targets' number and randomly select a specific number of checkerboards from the reference set for calibration (repeated 100 times independently). The mean reprojection error is used to represent the calibration accuracy. The results in Fig.~\ref{checkerboard_b} show that as the number of checkerboards increases, the calibration is more accurate and converged. It is likely that more checkerboards would increase the FoV coverage and feature points density and improve the effectiveness of the target-based method. However, it is labor-intensive to place many checkerboards uniformly and densely around the sensor and manually select the appropriate ones, which may be impossible in the field. The co-calibration method, on the contrary, employs dense LiDAR points as abundant, well-covered, and accurate features; and the elimination of artificial targets and human involvement enables an accurate, efficient, and field-friendly approach. Our co-calibration result yields a significantly improved performance on the same reference set, compared with the conventional method (orange and blue boxplot in Fig.~\ref{checkerboard_b}, respectively).

\subsubsection{Analysis of the Extrinsic Results}
\begin{figure}[ht]
    \centering {
        \subfloat
        {
            \includegraphics[width=0.3\linewidth]{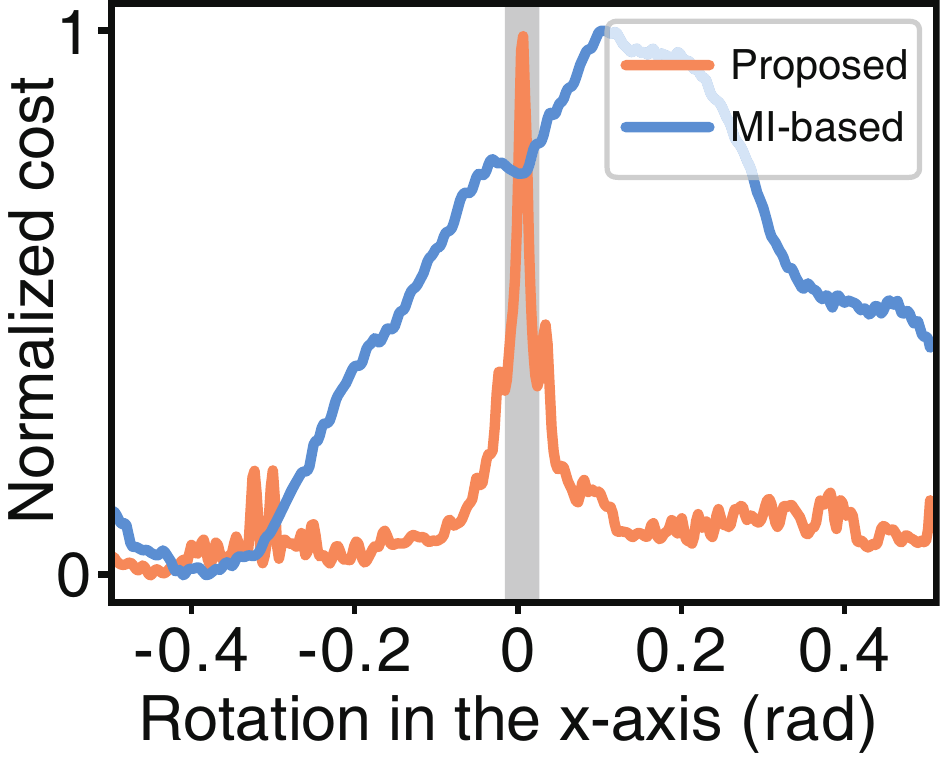}
        } %
        \subfloat
        {
            \includegraphics[width=0.3\linewidth]{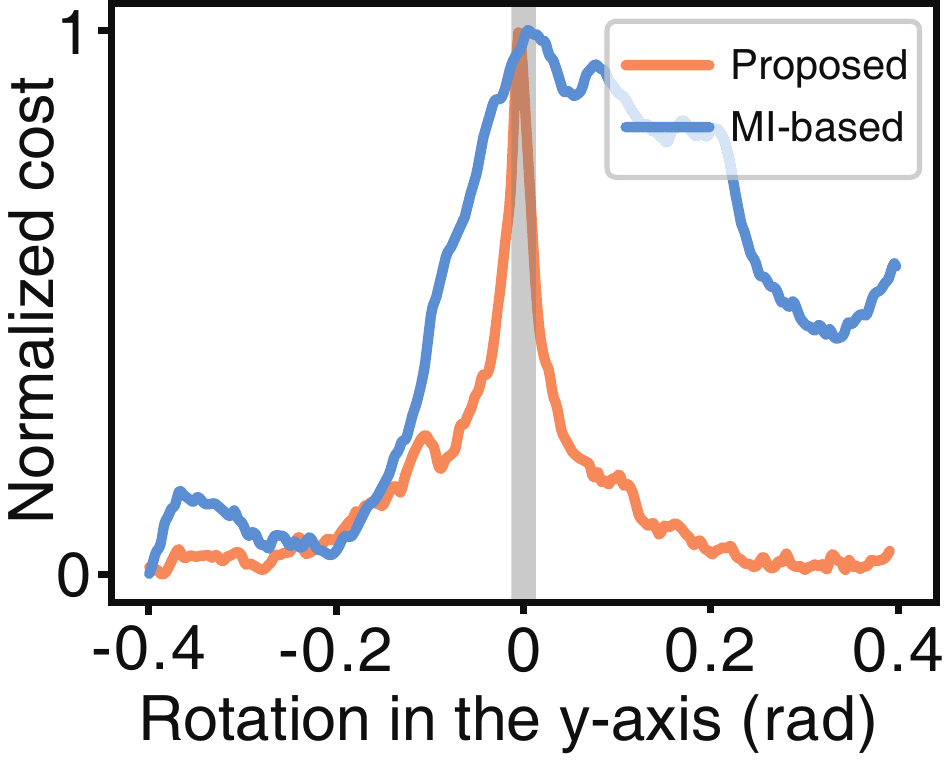}
        } % 
        \subfloat
        {
            \includegraphics[width=0.3\linewidth]{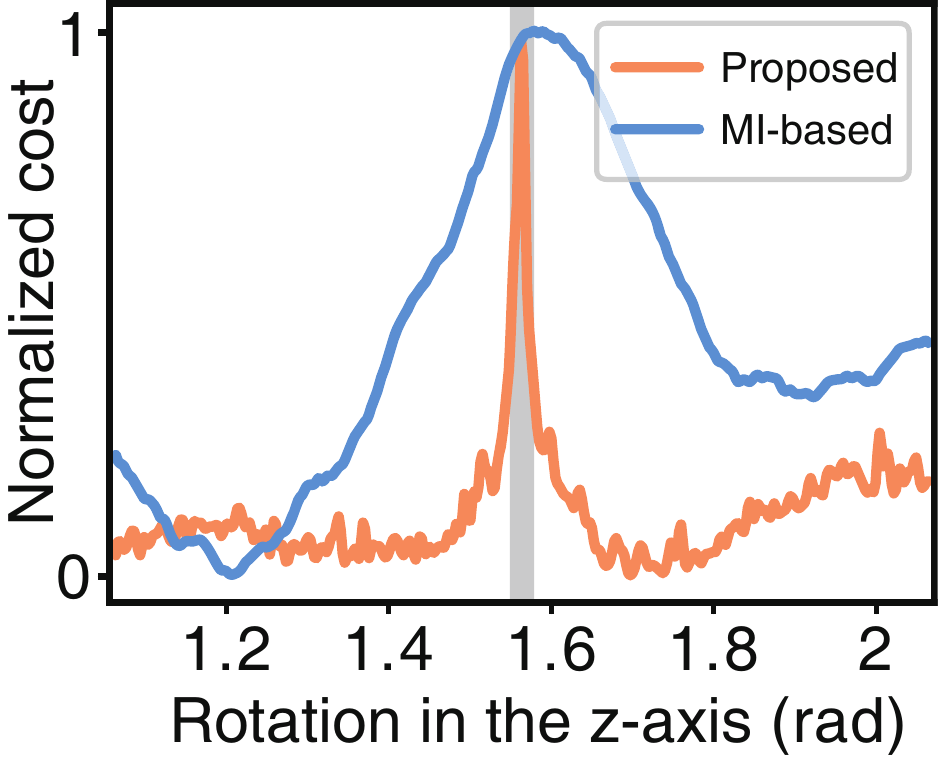}
        } % 
    } %
% 	\hspace{6pt}
	\vfill
    \centering {
        \subfloat
        {
            \includegraphics[width=0.3\linewidth]{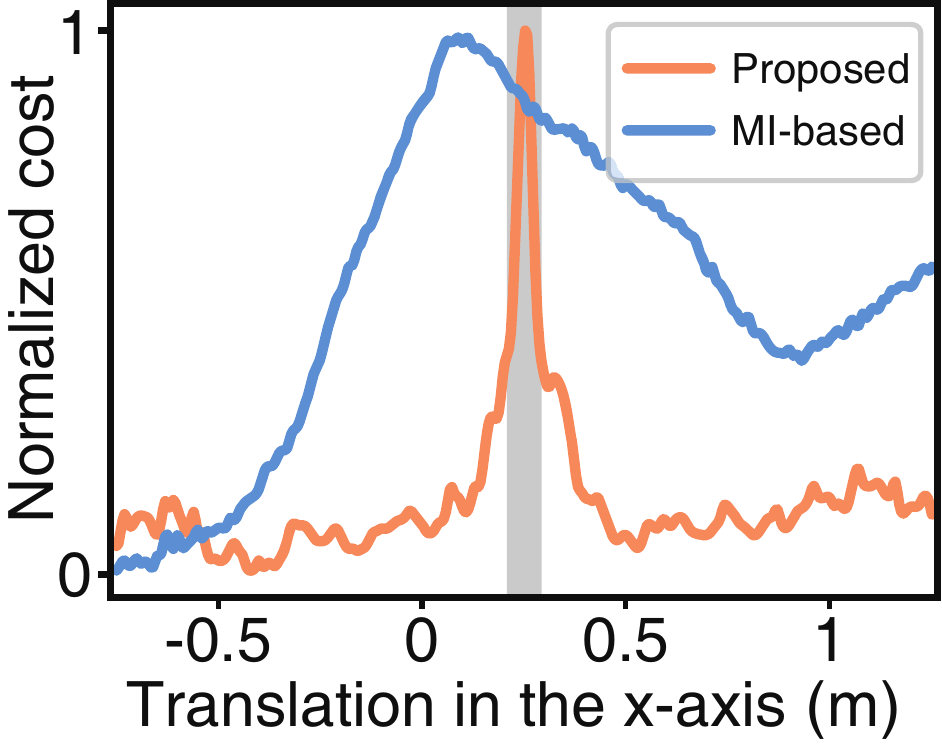}
        } %
        \subfloat
        {
            \includegraphics[width=0.3\linewidth]{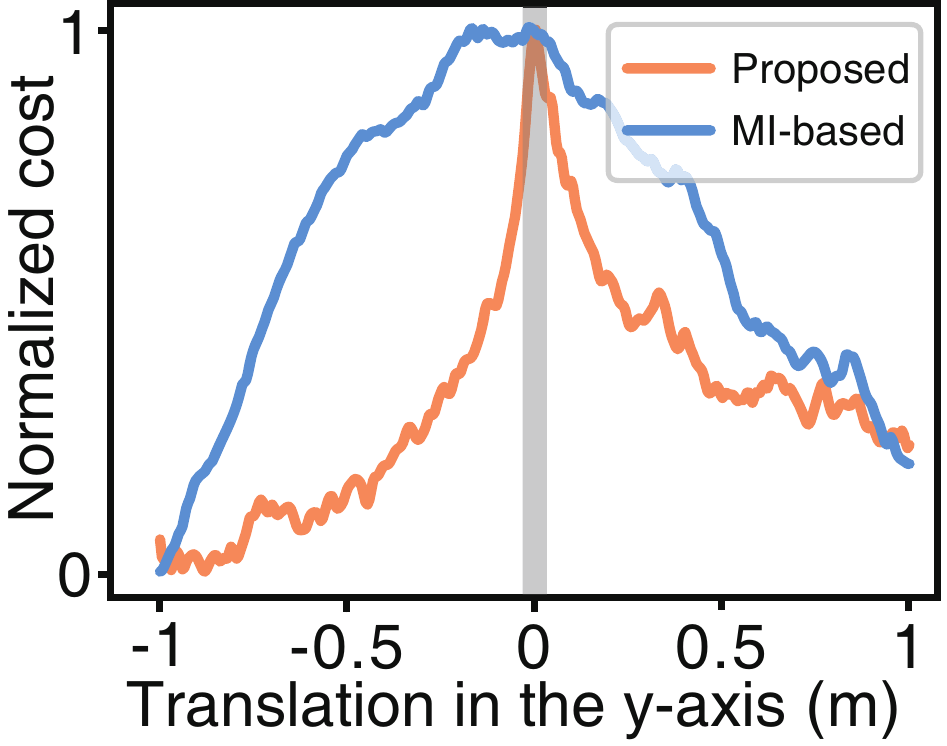}
        } %
        \subfloat
        {
            \includegraphics[width=0.3\linewidth]{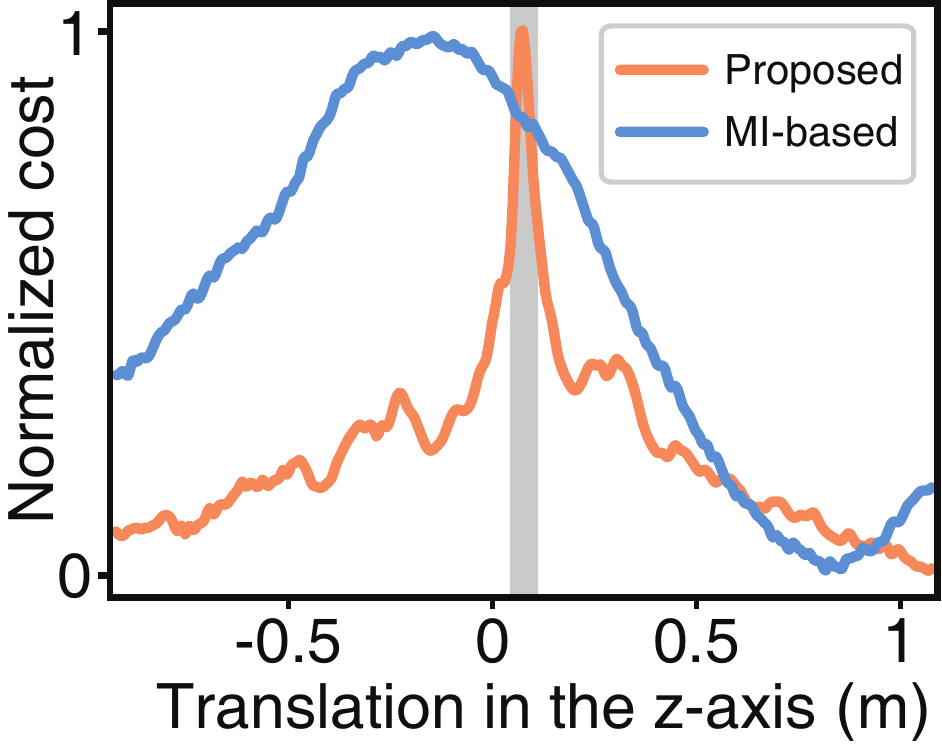}
        } % 
    } %
    \caption{Comparisons of the normalized cost function between the proposed method and the MI-based method. The optimal values should lie in the gray areas estimated based on manufacturing.}
    \label{cost}
\end{figure}

The mutual information (MI)-based extrinsic calibration method utilizes the fact that the reflectivity of LiDAR points and corresponding grayscale intensity values of camera pixels are correlated since both of them capture the spectral response of the object at light frequencies (LiDAR 905 nm, camera 400-800 nm), which are usually similar. These values are then used to calibrate the extrinsic parameters between the camera and LiDAR by maximizing the MI of the two distributions~\cite{mutual}. Fig.~\ref{cost} shows the comparisons of the two optimization methods demonstrating the normalized costs on different extrinsic parameters. The proposed co-calibration method shows a much more sensitive and reliable gradient in the cost function near the optimum than the MI-based method.

\vspace{-12pt}

\begin{figure}[ht]
    \centering{
        \subfloat[]{
           \includegraphics[width=0.85\linewidth]{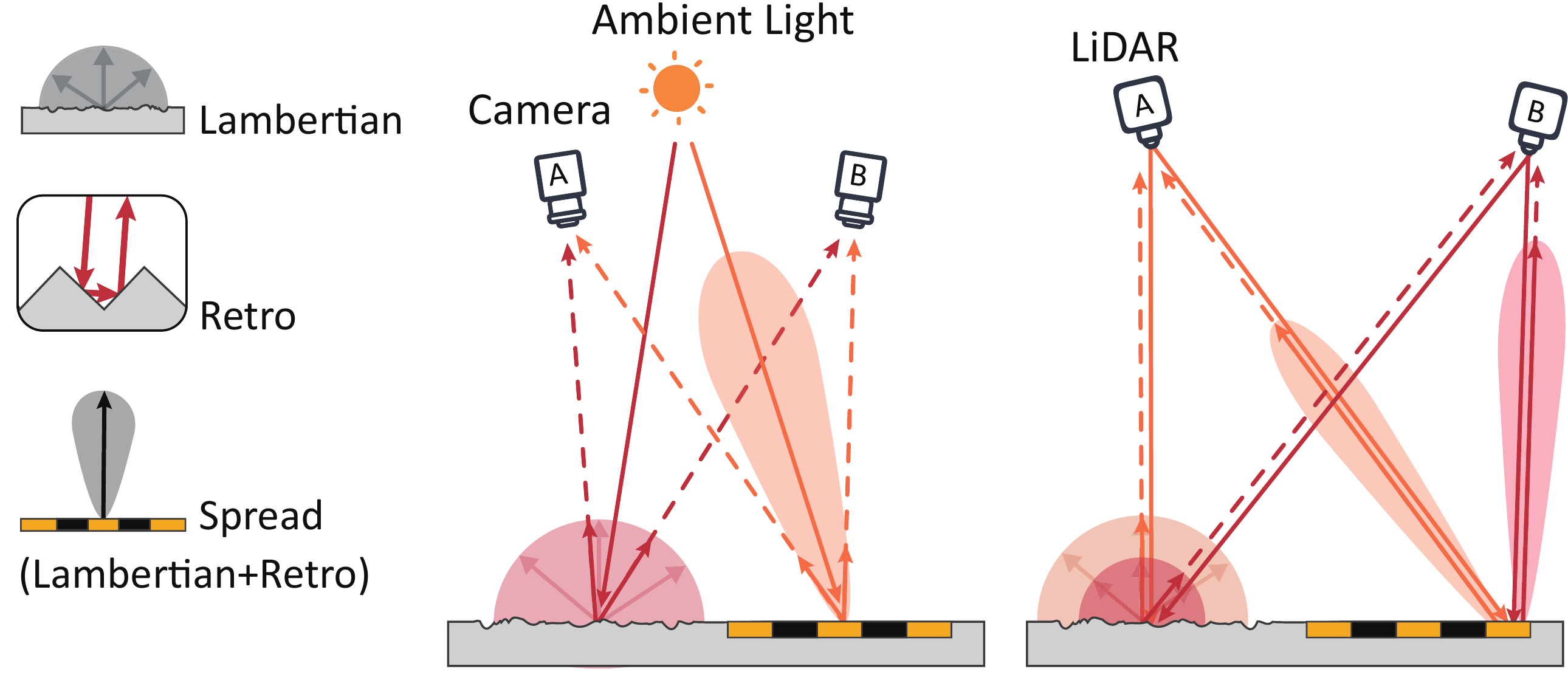}
           \label{reflection_a}
        } %
    } %
    \vfill
    {
    	\subfloat[]{
    	   \includegraphics[width=0.85\linewidth]{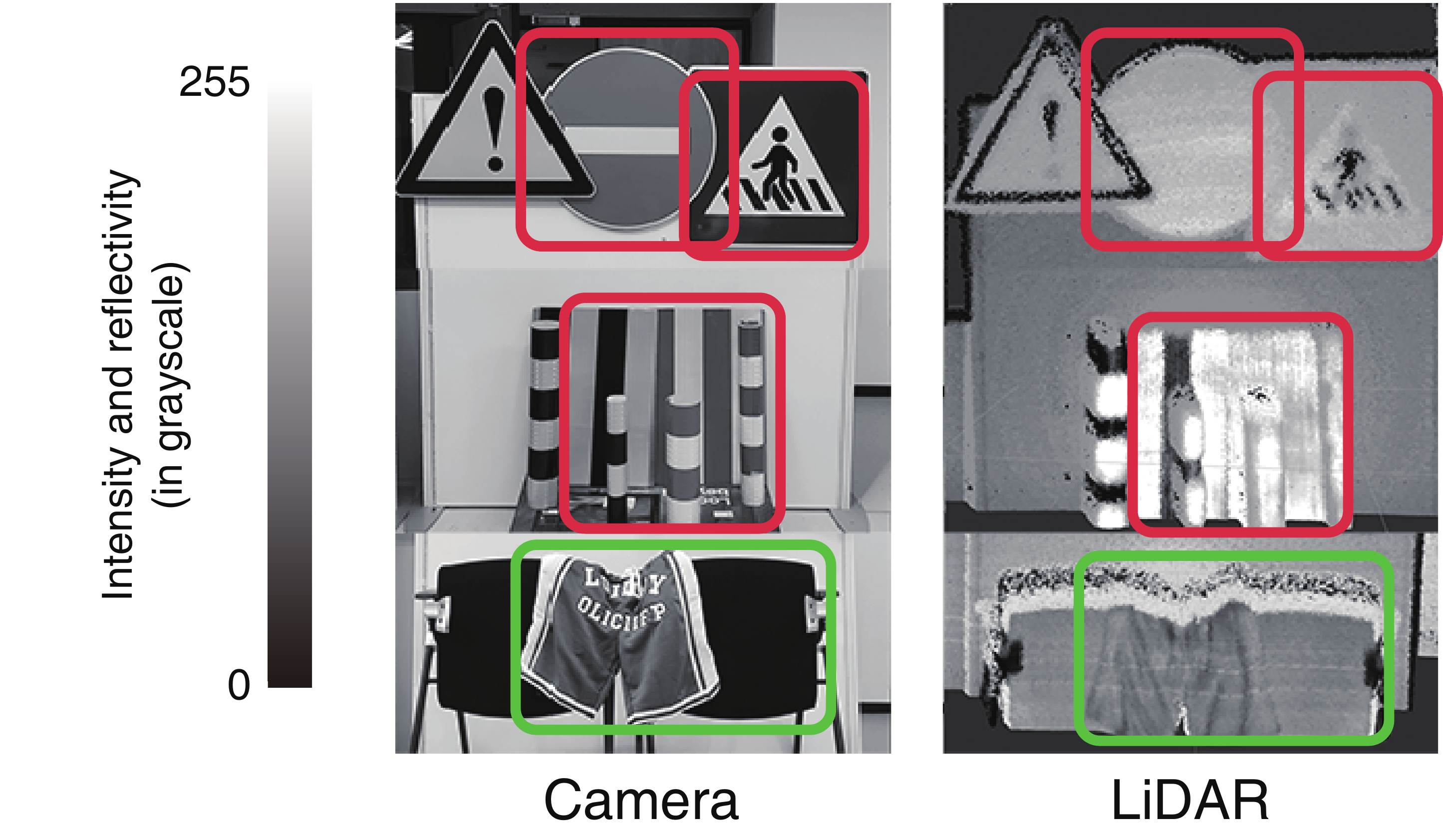}
              \label{reflection_b}
    	} %
	} %
    \caption{Analysis of the MI-based extrinsic calibration: (a) the types of reflection of the LiDAR and camera w.r.t. the rough surface and the retroreflective surface; (b) the inconsistent intensity cases between LiDAR and camera, including retroreflection cases (red boxes), and the special spectral reflectance cases (green boxes).}
    \label{reflection}
\end{figure}

The inaccurate calibration result of the MI-based method could be attributed mainly to three reasons: the lighting conditions, the surface reflection properties, and the spectral reflectance disagreement. The camera's light source $I_i$ is the external ambient lighting which does not change with the camera pose. On the contrary, LiDAR uses an active laser from the sensor and therefore differs significantly from the camera, as shown in Fig.\ref{reflection_a}. Besides the lighting, the surfaces of the objects are important. The detected intensity could be modeled as follows:
\begin{equation}
    I_r = K_d \cdot I_i \cdot f(\theta),
\end{equation}
where $I_r$ and $I_i$ indicate the reflection intensity and incident intensity, respectively, $K_d$ is the reflectance, and $f(\theta)$ describes the surface properties of the object with respect to incident angle $\theta$. For most objects, the surface is Lambertian (diffusive), and in that case, $f(\theta) = \cos{\theta}$. However, many surfaces do not follow this property, and it could be a specular reflection that the LiDAR does not collect any signal; or the retroreflection that the majority of the energy will be directed back toward the LiDAR itself and gives a strong intensity, such as those on traffic signs and warning stickers, which show a contrast difference in the LiDAR intensities from the camera intensities shown in the red boxes in Fig.~\ref{reflection_b}. Additionally, the spectral reflectance of objects at various light wavelengths could be different. For instance, materials composed of plant fibers show a large reflectance at around 905 nm, even those dyed in black colors. As a result, no contrast could be seen in LiDAR intensities of materials with different colors, as shown in green boxes in Fig.~\ref{reflection_b}. All three factors mentioned above could cause significant differences in intensity response from the LiDAR and the camera and reduce the applicability of the MI-based method.

\vspace{-24pt}
\subsection{Coarse-to-fine Hybrid Mapping Results}
\label{4.1}
\begin{figure}[ht]
    \centering
    \begin{minipage}[b]{0.45\linewidth}
        \centering
        \subfloat[]{
            \includegraphics[width=1\linewidth]{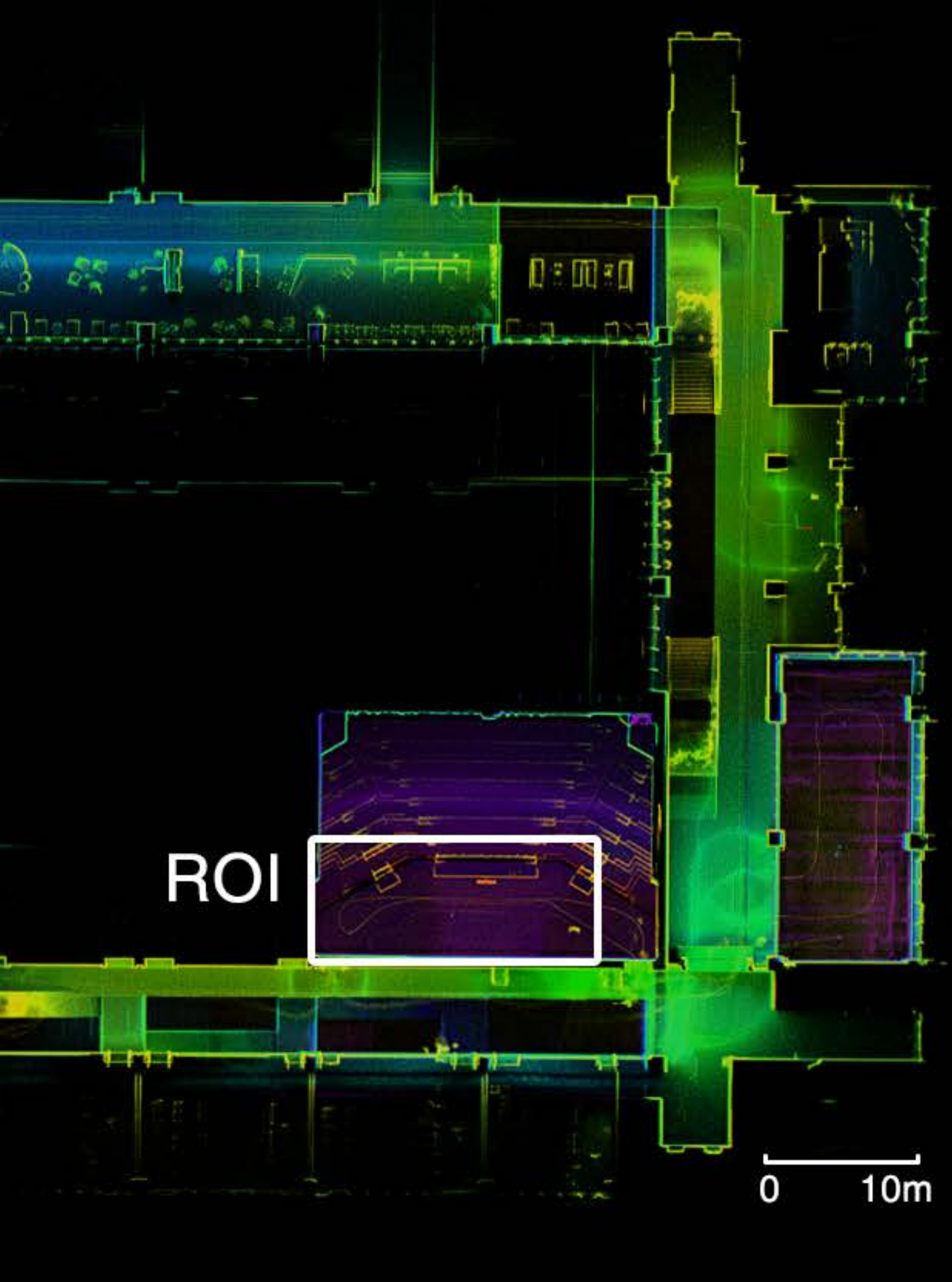}
            \label{hybrid_mapping_a}
        }
    \end{minipage} %\par
    \medskip
    \begin{minipage}[b]{0.46\linewidth}
        \centering
        \subfloat[]{
            \includegraphics[width=1\linewidth]{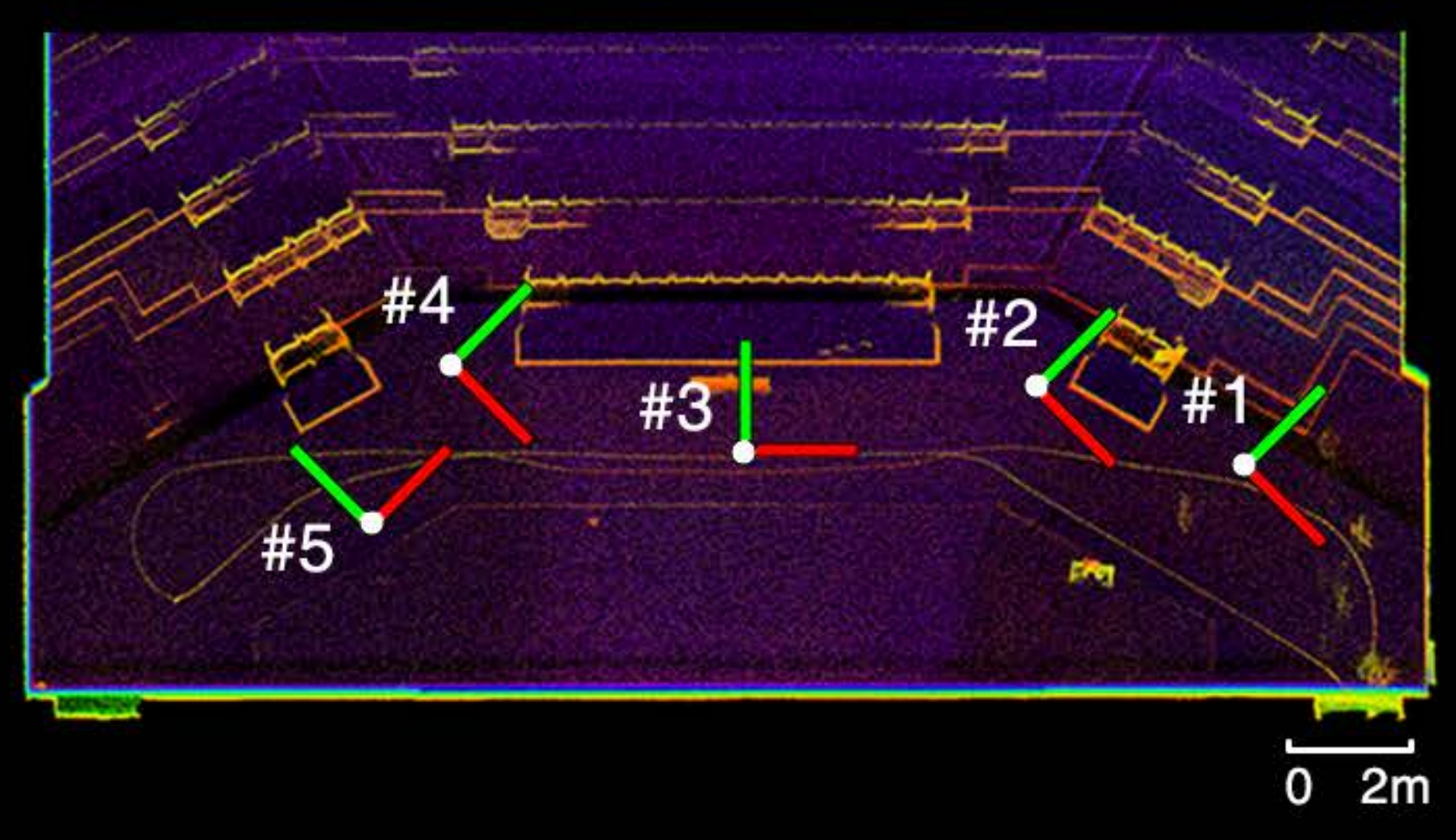}
            \label{hybrid_mapping_b}
        } %
        \vfill
        \subfloat[]{
            \includegraphics[width=1\linewidth]{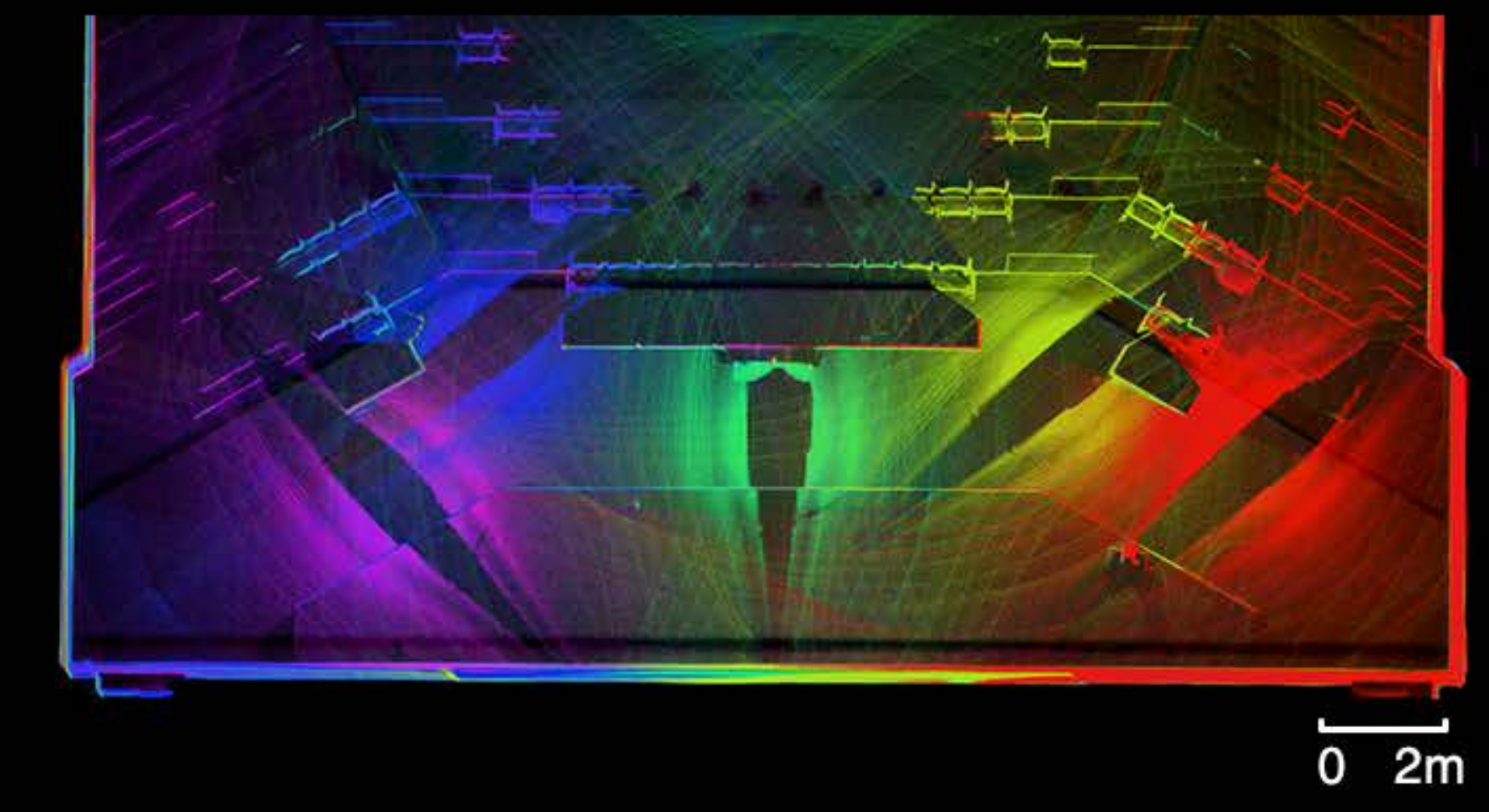}
            \label{hybrid_mapping_c}
        } %
    \end{minipage}
    \caption{Coarse-to-fine hybrid mapping: (a) odometry-based global coarse mapping; (b) coarse map of the selected ROI, with markers indicating the planned viewpoints; (c) fine map of the ROI, the color illustrates the scans from respective viewpoints.}
    \label{hybrid_mapping}
\end{figure}

The proposed coarse-to-fine hybrid mapping method is demonstrated in an academic building on the SUSTech campus. The global coarse map is generated by Fast-LIO in ten minutes, and the ROI is selected based on this global coarse map (Fig.~\ref{hybrid_mapping_a}). In this case, five viewpoints are properly planned in this ROI (Fig.~\ref{hybrid_mapping_b}), and perform stationary scanning for three minutes in each (Fig.~\ref{hybrid_mapping_c}).
\begin{table}[h]
% 注意: Minipage之间不能有空行，否则会自动换行
\makeatletter\def\@captype{table}\makeatother
\centering
\setlength\tabcolsep{3pt}
\renewcommand\arraystretch{1.1}
\vspace{-6pt}
\caption{Specs Comparison of Current Mapping Systems}
\label{comparison_table}

\scalebox{0.9}{
\begin{tabular}{|c|c|c|}
% \hline
\toprule  %添加表格头部粗线
& \textbf{Proposed} & \textbf{\makecell[c]{\#1 FARO Focus\\Premium 150}}\\
% \midrule  %添加表格中横线
\hline
\textbf{Type} & \makecell[c]{Hybrid Mapping} & TLS\\
\hline
\textbf{FoV} & $360\degree \times 300\degree$ & $360\degree \times 300\degree$\\
\hline
\textbf{Range} & 0.1-40 m & 0.5-150 m\\
\hline
\textbf{PPS} & 200,000 pts/s & 2,000,000 pts/s\\
\hline
\textbf{Precision} & \makecell[c]{$\sim$ 40 mm (coarse)\\$\sim$ 20 mm (fine)} & $\sim$ 1mm~\cite{faro}\\
\hline
\textbf{Accuracy} & \makecell[c]{$\sim$ 10 mm (coarse)\\$\sim$ 2 mm (fine)} & $\sim$ 1mm~\cite{faro}\\
\hline
\textbf{Registration} & \makecell[c]{Odometry+Optimization} & \makecell[c]{Optimization}\\
\hline
\textbf{Work Manner} & Mobile Robot & Manual (tripod)\\
\hline
\textbf{\makecell[c]{Viewpoints Planning}} & Coarse map-based & Intuition-based\\
\hline
\textbf{Vision} & 1-omni camera & 1-camera\\
\bottomrule %添加表格底部粗线
% \hline

% \hline
\toprule
\textbf{\makecell[c]{\#2 LEICA\\BLK360}} & \textbf{\makecell[c]{\#3 LEICA\\BLK2GO}} & \textbf{\makecell[c]{\#4 NavVis\\VLX}}\\
\hline
TLS & MLS & MLS\\
\hline
$360\degree \times 300\degree$ & $360\degree \times 270\degree$ & $360\degree \times 30\degree (\times2)$\\
\hline
0.5-45 m & 0.5-25 m & 0.9-100 m\\
\hline
680,000 pts/s & 420,000 pts/s & 300,000 pts/s ($\times2$)\\
\hline
$\sim$ 20 mm~\cite{comparison} & $\sim$ 20 mm~\cite{comparison} & 15-50 mm(walls, 80.5\%)~\cite{navvis_accuracy}\\
\hline
$\sim$ 1 mm~\cite{comparison} & $\sim$ 30 mm~\cite{comparison} & 15-50 mm(beams, 98.2\%)~\cite{navvis_accuracy}\\
\hline
Optimization & Odometry/SLAM & Odometry/SLAM\\
\hline
Manual (tripod) & Manual (handheld) & Manual (backpack)\\
\hline
Intuition-based & No need & No need\\
\hline
3-camera & 3-camera & 4-camera\\
% \hline
\bottomrule
\end{tabular}
} %
\vspace{-12pt}
\end{table}

\begin{figure}[ht]
    \centering
    \subfloat[]{
    \includegraphics[width=0.9\linewidth]{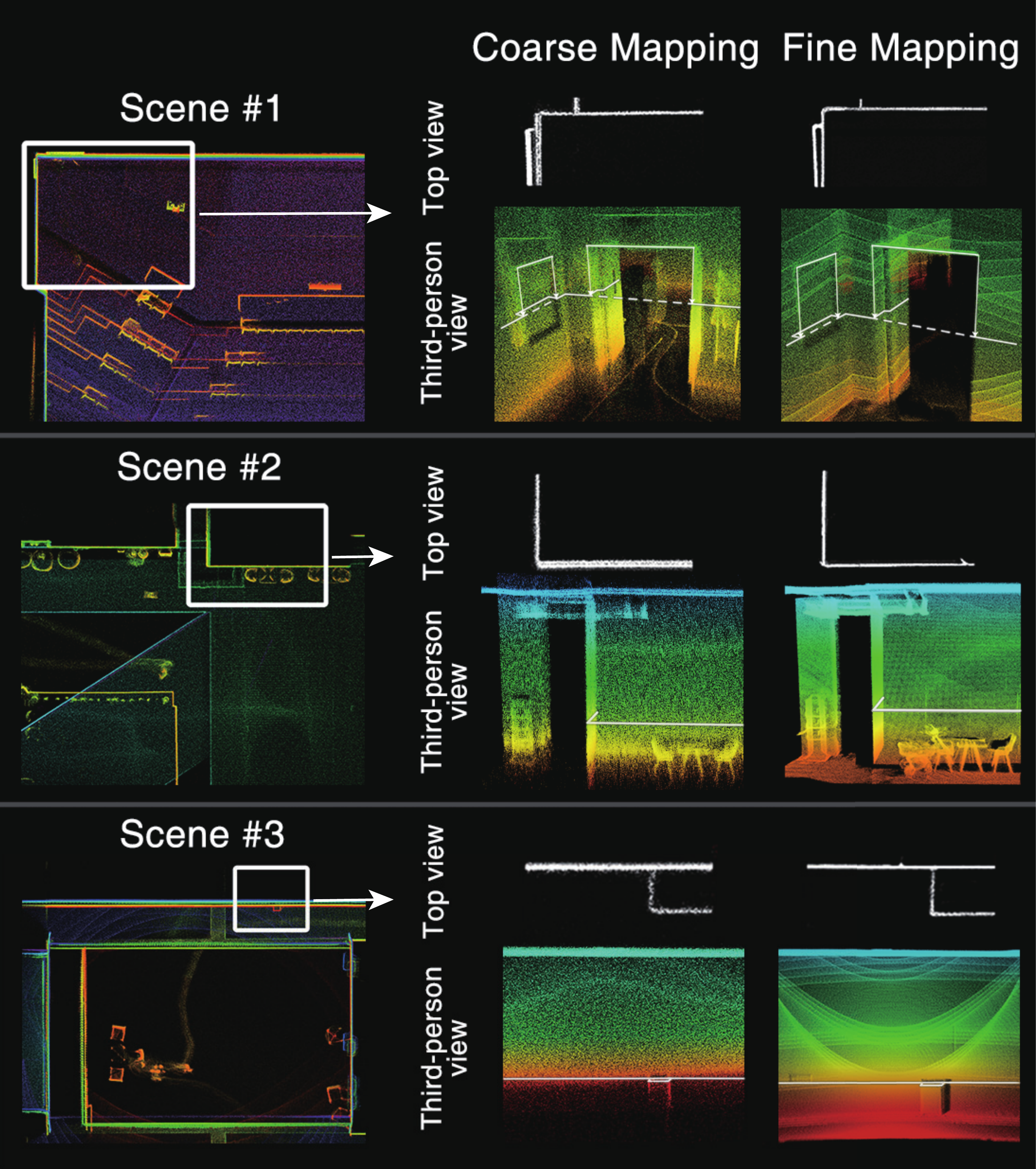}
        \label{precision_a}
    } %
    \vfill
    \centering
    \subfloat[]{    \includegraphics[width=0.45\linewidth]{pics/precision-b.pdf}
    \label{precision_b}
    } %
    \subfloat[]{\includegraphics[width=0.45\linewidth]{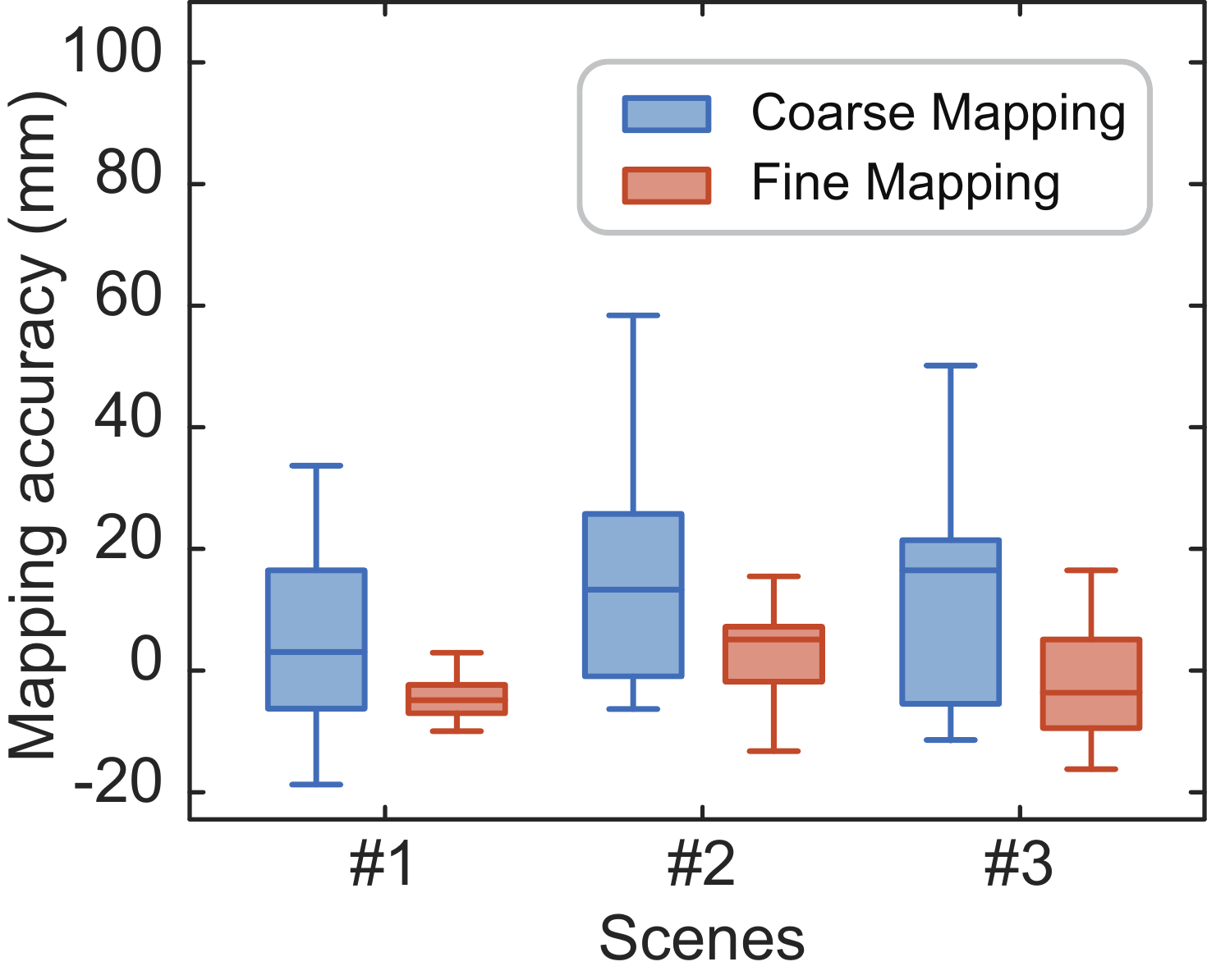}
    \label{accuracy}
    } %
    \vfill
    \subfloat[]{
        \label{colorized_mapping_a}
        \includegraphics[width=0.45\linewidth]{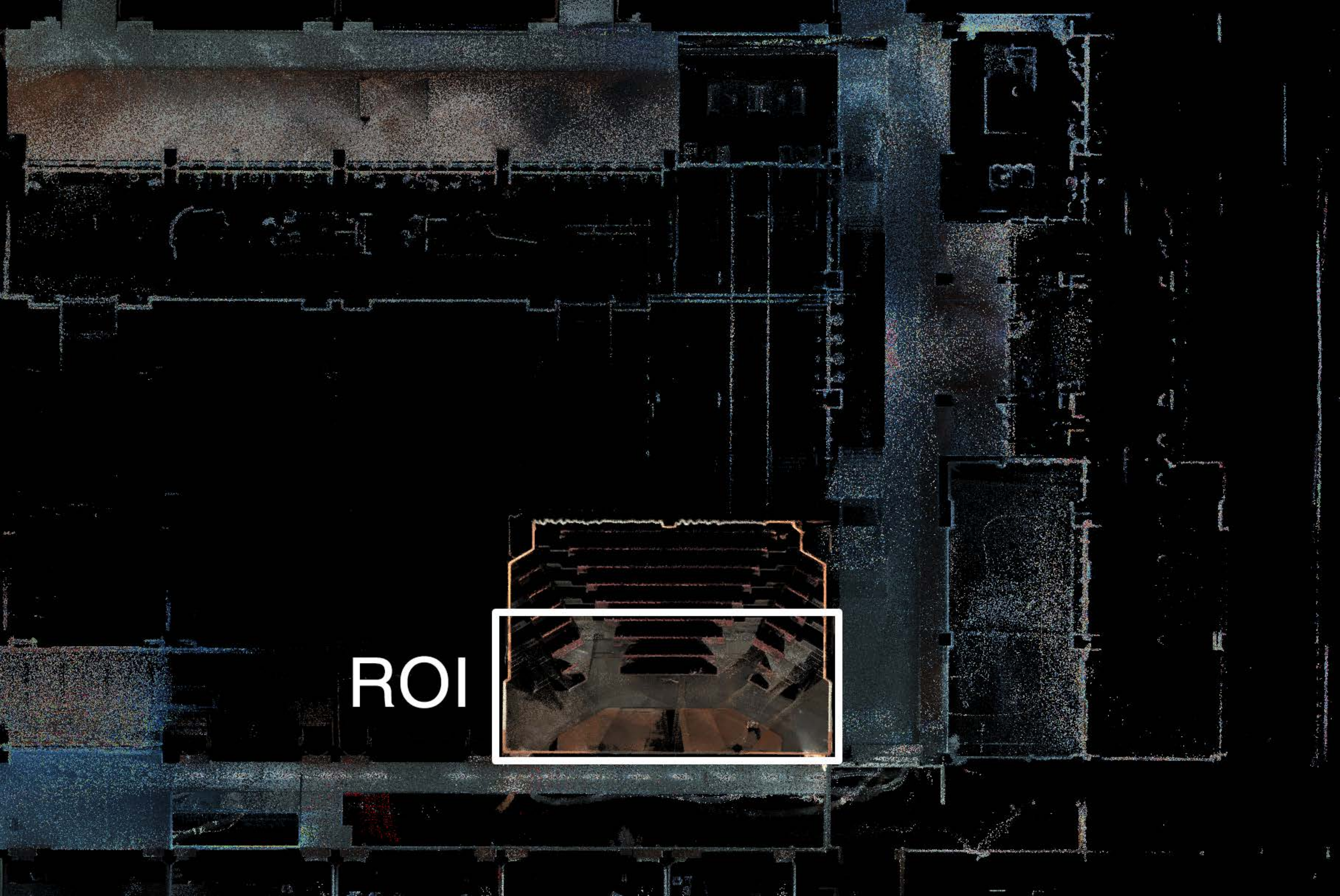} 
    } %
    \subfloat[]{
        \label{colorized_mapping_b}
        \includegraphics[width=0.45\linewidth]{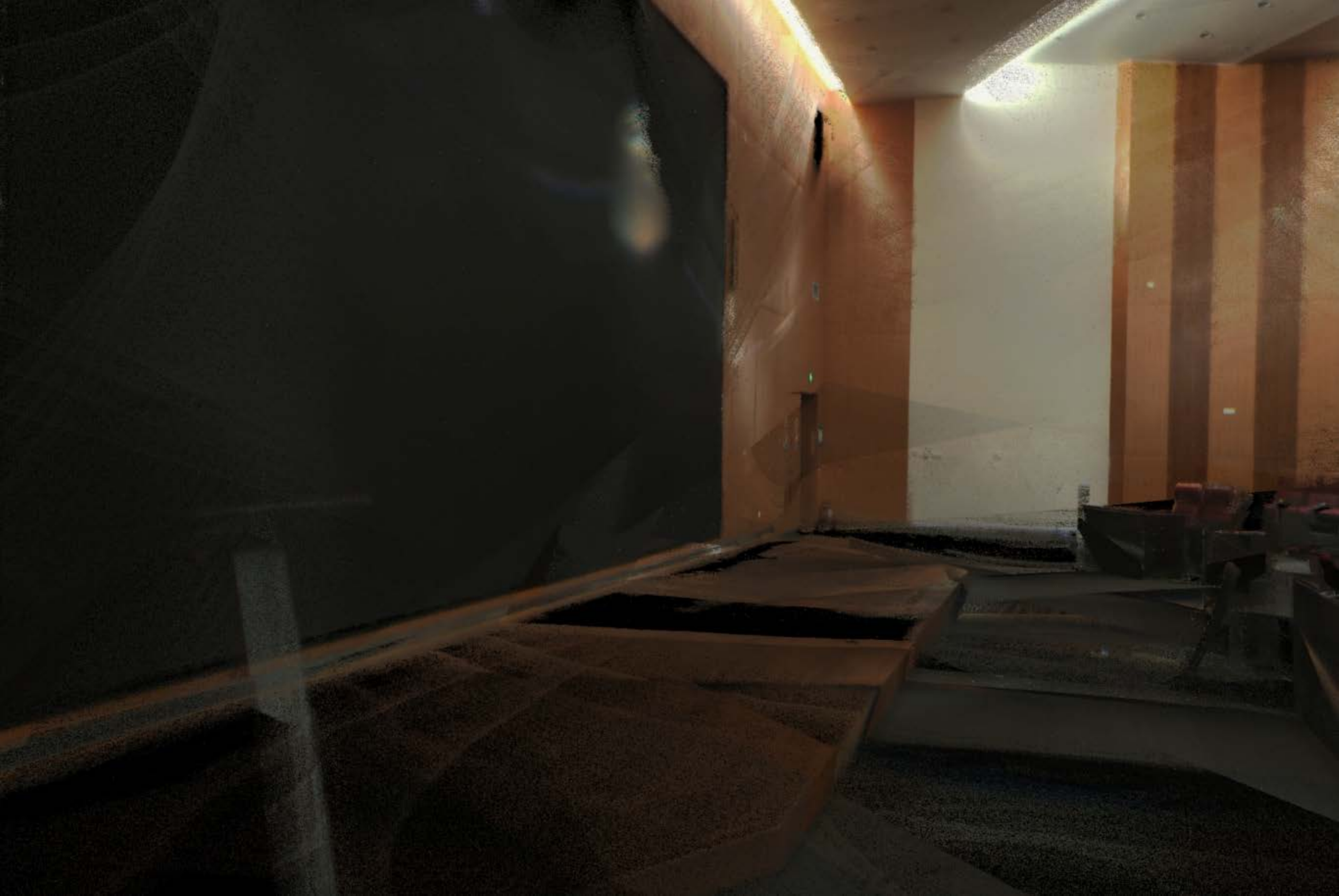}
    } %
    \caption{Comparison of coarse and fine mapping: (a) coarse and fine maps in three scenes (scene \#1 is from Fig.~10b, scene \#2 and \#3 are new). The left column shows the large-scale coarse map, and the right column shows the zoomed-in coarse and fine map in top view (to visualize wall thickness) and third person view (to visualize scene); (b) mapping precision from the three scenes; (c) mapping accuracy from the three scenes; (d) top view of the colorized fine map; (e) third-person view of the colorized ROI.}
    \label{precision}
\vspace{-12pt}
\end{figure}

Plane thickness could be used as a quantitative metric for precision evaluation and comparison between coarse and fine mapping. Local planes with a small third eigenvalue $\lambda_3$ are selected by diagonalizing the covariance matrix. Assuming the points along the plane's normal direction follow the Gaussian distribution (corresponding to the third eigenvalue $\lambda_3$ with the normal direction of the plane defined by its eigenvector), we could set the thickness of the plane as $4\sqrt{\lambda_3}$. The coarse and fine maps of the three different scenes are shown in Fig.~\ref{precision_a}, whereas the zoomed views show the point cloud quality with the top view of the selected planes to demonstrate the mapping quality. The quantitative evaluations of the plane thickness (the mapping precision) in these scenes are summarized in Fig.~\ref{precision_b}. Besides precision (spread of data), accuracy (correctness) is also important to examine. Fig.~\ref{accuracy} illustrates the measurement accuracy (compared to results from a TLS system, which we regard as ground truth). It is evident that both the precision and accuracy of fine mapping outperform coarse mapping. Although odometry-based coarse mapping has good performances in best-case scenarios, it could be significantly improved by fine mapping in the average values and worse-case scenarios, which are the main concerns of the surveying and mapping industry. 

With the accurate co-calibration results, LiDAR points can be colorized from the image information through the transformation in Eqn.~\ref{projection1} and Eqn.~\ref{projection2}. Fig.~\ref{colorized_mapping_a} shows the colorized hybrid mapping, and Fig.~\ref{colorized_mapping_b} illustrates the fine mapping of the zoomed-in ROI. The coarse-to-fine map with great precision and accurate colorization pave the way for higher precision with a single unified setup and workflow. It benefits industries requiring both efficiency and accuracy, such as construction automation and building inspection.

Lastly, a detailed comparison of the proposed system with the current widely used TLS and MLS systems (shown in Fig.~\ref{system_a}) is made in Table~\ref{comparison_table}, where several key parameters are listed. The most crucial difference is that the proposed system integrates two working modes in a single streamlined workflow, ensuring overall mapping efficiency and precision/accuracy. All other systems are either TLS which only works in stationary mode, or MLS in mobile mode. Due to this capability, it is the first robotic system that allows automatic viewpoint planning instead of human intuition-based viewpoints selection. In addition, the mobile robot could navigate itself with overall good localization and provide good initial states for fine map optimization. The mapping precision and accuracy of the proposed system are also compared with these systems \cite{faro, comparison, navvis_accuracy}. The proposed system achieves performance close to the LEICA TLS but allows mobility as MLS, agreeing with the purpose of the system.

\section{Conclusion}
This paper proposed a coarse-to-fine hybrid 3D mapping robotic system based on an omnidirectional camera and a non-repetitive Livox LiDAR. A hybrid mapping approach with both odometry-based and stationary mapping modes is integrated into one mobile mapping robot, achieving a streamlined and automated mapping workflow with the assurance of efficiency and mapping precision and accuracy. Meanwhile, the proposed automatic and targetless co-calibration method provides accurate parameters to generate colorized mapping. Specifically, the calibration is based on edges extracted from camera images and LiDAR reflectivity, and the result is compared with the mutual-information-based calibration method, which was under-performing possibly due to varied reflection nature in light sources, surface reflection properties, and the spectral reflectance disagreement in the MI-based method. In future work, more complicated planning strategies could be developed to further optimize both the objectives of scanning time and spatial coverage. We believe this new automated mapping robot will open up a new horizon for surveying and inspection robotics.

\bibliographystyle{IEEEtran}
\bibliography{ref}
\end{document}